\documentclass[nonblindrev]{informs3_custom} 

\DoubleSpacedXI 

\usepackage[draft]{hyperref}
\RequirePackage{fix-cm}
\usepackage{tikz}
  \usetikzlibrary{bayesnet}
  \usetikzlibrary{arrows} 

\usepackage{xspace}
	\newcommand\ie{i.\,e.\xspace}
	\newcommand\eg{e.\,g.\xspace}

  \newcommand{\abs}[1]{\left\lvert #1 \right\rvert}

	\newcommand*\dif{\mathop{}\!\mathrm{d}}
	\newcommand{\dd}{\mathop{}\!\mathrm{d}}

	\usepackage{siunitx}
    \def\sym#1{\ifmmode^{#1}\else\(^{#1}\)\fi}
    \DeclareSIUnit\eur{\officialeuro}
    \DeclareSIUnit\M{M}
    \DeclareSIUnit\k{k}
	
	\usepackage[%
  sort&compress
]{cleveref}
  \crefname{chapter}{section}{sections}
	\Crefname{chapter}{Section}{Sections}

\usepackage{etoolbox}
\robustify\bfseries

\usepackage{url}

\usepackage{isomath}
\usepackage{bm}

\usepackage{colortbl}
\usepackage{tabularx}
\usepackage{booktabs}
\usepackage{lscape}
\usepackage{collcell}
\usepackage{subfig}

\usepackage{float}
\usepackage{rotating}

\usepackage{array}
\newcolumntype{L}[1]{>{\raggedright\let\newline\\\arraybackslash\hspace{0pt}}p{#1}}
\newcolumntype{C}[1]{>{\centering\let\newline\\\arraybackslash\hspace{0pt}}p{#1}}
\newcolumntype{R}[1]{>{\raggedleft\let\newline\\\arraybackslash\hspace{0pt}}p{#1}}

\usepackage{csquotes}
\usepackage{amsmath}
\usepackage{amssymb}
\usepackage{pifont}

\newcommand{\lcellt}[2][l]{%
  \begin{tabular}[t]{@{}#1@{}}#2\end{tabular}}

\usepackage[
    colorinlistoftodos,
    textsize=footnotesize,
        ]{todonotes}

\newcommand{\REV}[1]{{\color{black}#1}}

\newcommand{\LATER}[1]{}
\newcommand{\FINAL}[1]{}
        
\makeatletter
    \renewcommand{\fps@figure}{H}         
    \renewcommand{\fps@table}{H}         
\makeatother 

\usepackage{listings}
\usepackage{lstbayes}

\usepackage{float}
\usepackage{rotating}

\makeatletter
\newcolumntype{B}[3]{>{\boldmath\DC@{#1}{#2}{#3}}c<{\DC@end}}
\makeatother

\usepackage{natbib}
 \bibpunct[, ]{(}{)}{,}{a}{}{,}%
 \def\bibfont{\small}%
\TheoremsNumberedThrough     

\EquationsNumberedThrough    

\begin{document}



\RUNTITLE{Predicting Customer Satisfaction}

\TITLE{Beyond Means: A Dynamic Framework for Predicting Customer Satisfaction}

\ARTICLEAUTHORS{%
\AUTHOR{Christof Naumzik}
\AFF{ETH Zurich, Switzerland, \EMAIL{christof.naumzik@googlemail.com}, \URL{}}
\AUTHOR{Abdurahman Maarouf\footnote{\footnotesize Corresponding author, Postal address: LMU Munich - Institute of AI in Management, Geschwister-Scholl-Platz 1, 80539 Munich, Germany}}
\AFF{LMU Munich \& Munich Center for Machine Learning (MCML), Germany, \EMAIL{a.maarouf@lmu.de}, \URL{}}
\AUTHOR{Stefan Feuerriegel}
\AFF{LMU Munich \& Munich Center for Machine Learning (MCML), Germany, \EMAIL{feuerriegel@lmu.de}, \URL{}}
\AUTHOR{Markus Weinmann}
\AFF{University of Cologne, Germany, \EMAIL{weinmann@wiso.uni-koeln.de}, \URL{}}
} 

\ABSTRACT{%
Online ratings influence customer decision-making, yet standard aggregation methods---such as the sample mean---fail to adapt to quality changes over time and ignore review heterogeneity (e.g., review sentiment, a review's helpfulness).
\REV{To address these challenges, we demonstrate the value of using the Gaussian process (GP) framework for rating aggregation.} Specifically, we present a tailored GP model that captures the dynamics of ratings over time while additionally accounting for review heterogeneity. Based on 121,123 ratings from Yelp, we compare the predictive power of different rating aggregation methods in predicting future ratings, thereby finding that the GP model is considerably more accurate and reduces the mean absolute error by 10.2\% compared to the sample mean. Our findings have important implications for marketing practitioners and customers. By moving beyond means, designers of online reputation systems can display more informative and adaptive aggregated rating scores that are accurate signals of expected customer satisfaction.
}


\HISTORY{}

\maketitle

\renewcommand\appendixname{Appendix}



\sloppy
\raggedbottom

\clearpage

\section{Introduction}


Online ratings\footnote{Typically in the form of star ratings and/or numerical scores.} play a vital role in many online interactions \citep[\eg,][]{Godes.2004,Moe.2011,Moe.2012}, not only to reduce uncertainty \citep{Pavlou.2004} but also to build trust \citep{Resnick.2000}. Will a seller deliver a product? Is this restaurant a high-quality restaurant? Will this app work properly? Hence, online ratings are used by many online platforms, including e-commerce platforms (\eg, Amazon), travel platforms (\eg, TripAdvisor), and app stores (\eg, Google Play). These ratings help customers learn more about a product or service so that they can draw conclusions about its quality \citep[\eg,][]{Chevalier.2006,Chintagunta.2010}. In the best case, ratings allow a customer to foresee how satisfied he or she will be with a product or service \citep{He.2013}. The data for inferring such expectations of satisfaction are available on many review platforms and, in theory, are directly accessible by customers. However, due to people's limited cognitive abilities, it is difficult for them to process large quantities of data, such as rating sequences \citep{Pope.2009}. One solution is for customers to base their decisions upon simple metrics such as an overall rating.


Recent studies show that aggregated ratings are essential for customers. For example, 83\,\% of customers searching for accommodations rely on aggregated ratings.\footnote{\SingleSpacedXI\footnotesize See \url{https://www.tripadvisor.com/business/insights/resources/bubble-rating}, accessed March 12, 2025.} Such aggregated ratings (often also referred to as ``overall ratings'') are typically user-independent and free of personalization \citep{Dai.2018}, either because websites have no access to user-specific information or personalization is unwanted. This is true for many comparison websites, such as IMDb, Rotten Tomatoes, Google Maps, Yelp, and TripAdvisor. These websites only have access to consumption data from their registered users, that is, actual review contributors. For Yelp, it is estimated that only 0.67\% of users are registered contributors, while the remaining users are visitors.\footnote{\SingleSpacedXI\footnotesize See \url{https://askwonder.com/research/percentage-yelp-users-mainly-write-reviews-restaurants-whrwmul7u}, accessed March 12, 2025.} Moreover, the main selling point of Yelp is the comparison of products or services. As a result, users are not required to sign in to use the website. Forcing users to register on the website would probably be counterproductive and decrease activity, as users might not want to sign up due to privacy reasons. Hence, these websites provide an overall score that is the same across the entire user base. 

\begin{figure}[htb]
\OneAndAHalfSpacedXI
\centering
\caption{Screenshot showing aggregated ratings (red boxes) on different rating platforms (Yelp, IMDb, Google Maps, Amazon). Note: the same ratings are shown for \emph{all} website visitors (\ie, no personalization). \label{fig:screenshot}}
\begin{tabular}{ccc}
\begin{minipage}[t]{.3\linewidth}
\centering
(a) Yelp\\
\fbox{\includegraphics[width=\textwidth]{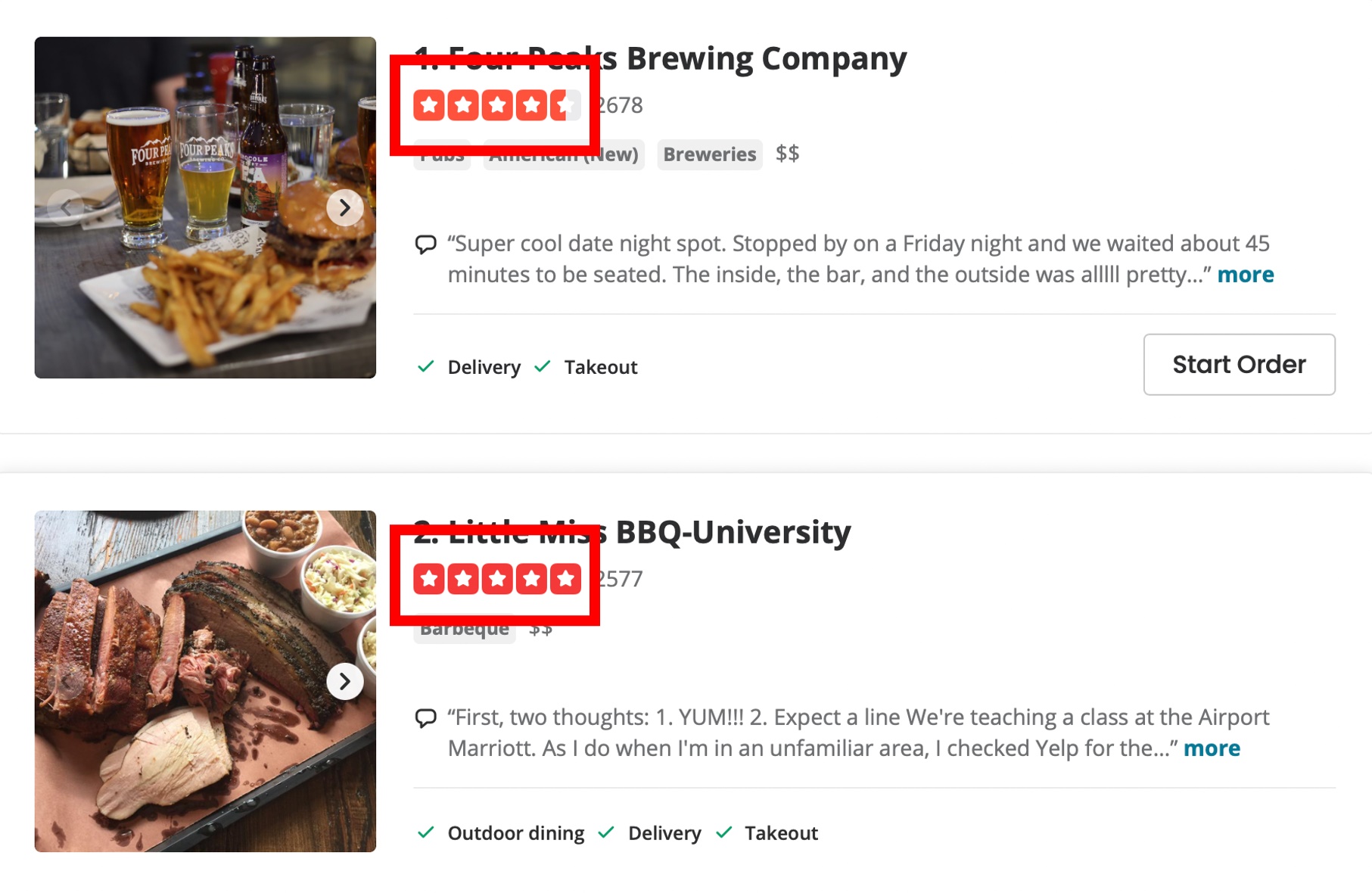}}
\end{minipage}
& \quad &
\begin{minipage}[t]{.5\linewidth}
\centering
(b) IMDb\\
\fbox{\includegraphics[width=\textwidth]{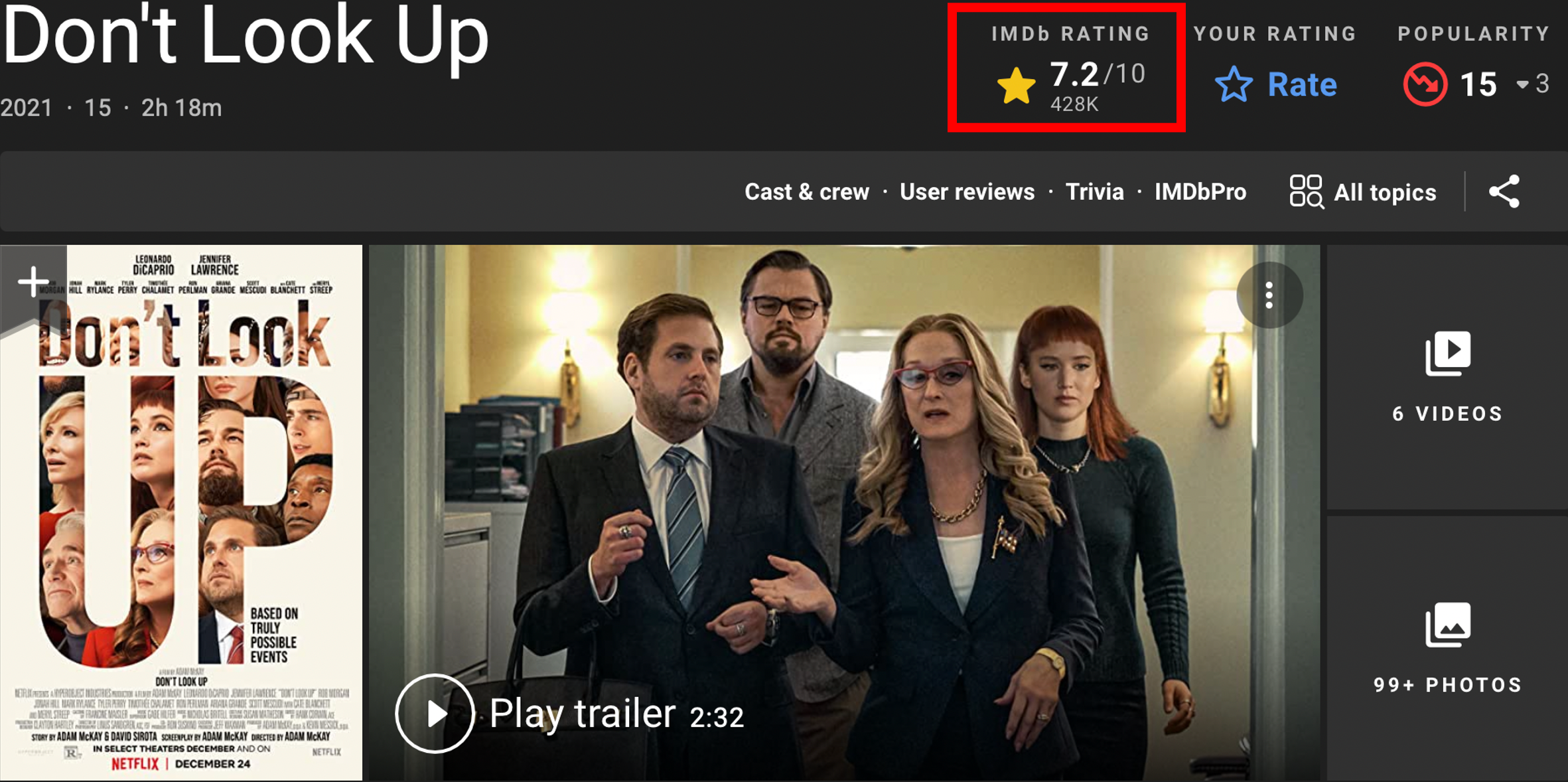}}
\end{minipage} \vspace{0.3cm} \\
\begin{minipage}[t]{.3\linewidth}
\centering
(c) Google Maps\\
\fbox{\includegraphics[width=\textwidth]{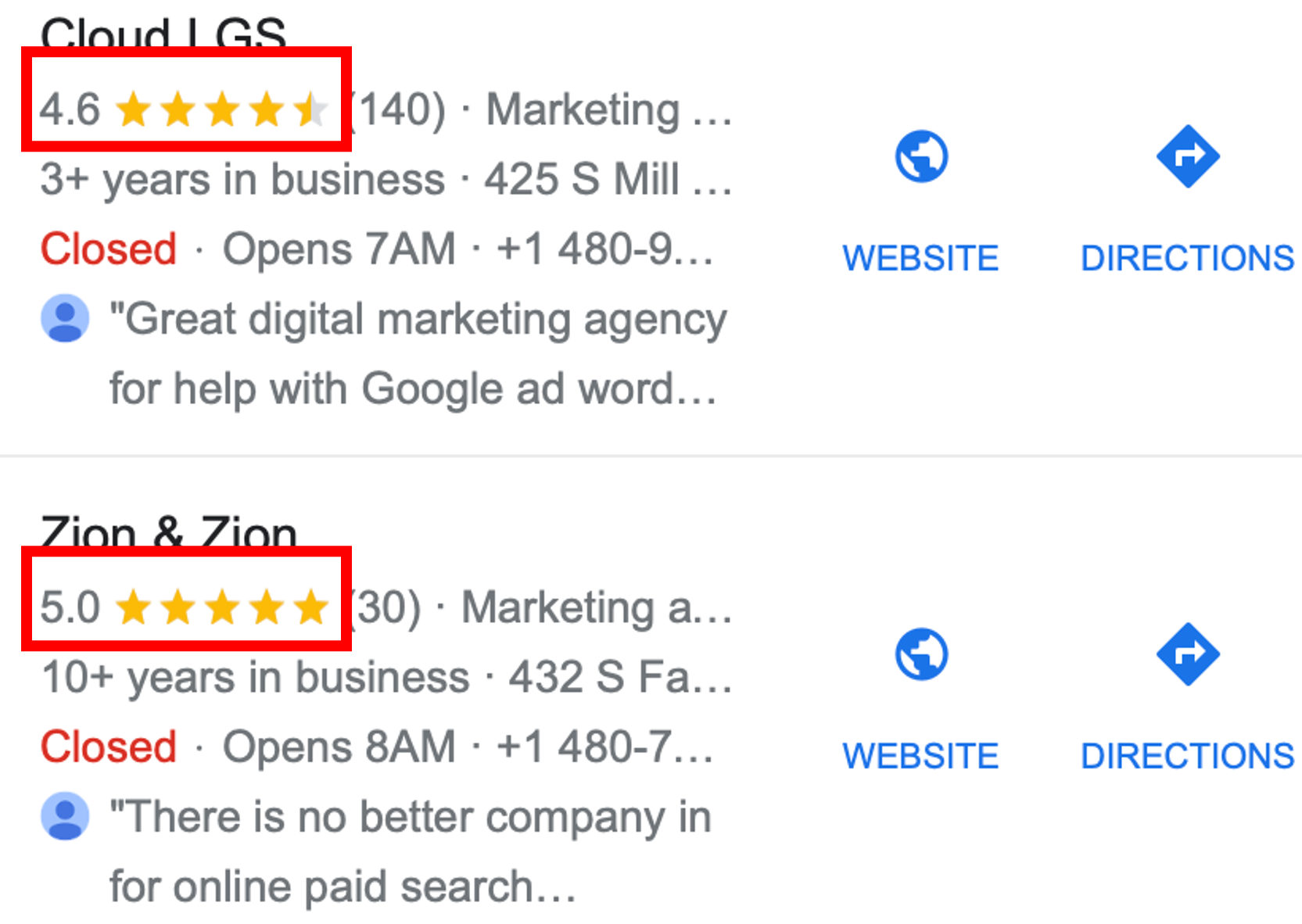}}
\end{minipage}
& \quad &
\begin{minipage}[t]{.5\linewidth}
\centering
(d) Amazon \\
\fbox{\includegraphics[width=\textwidth]{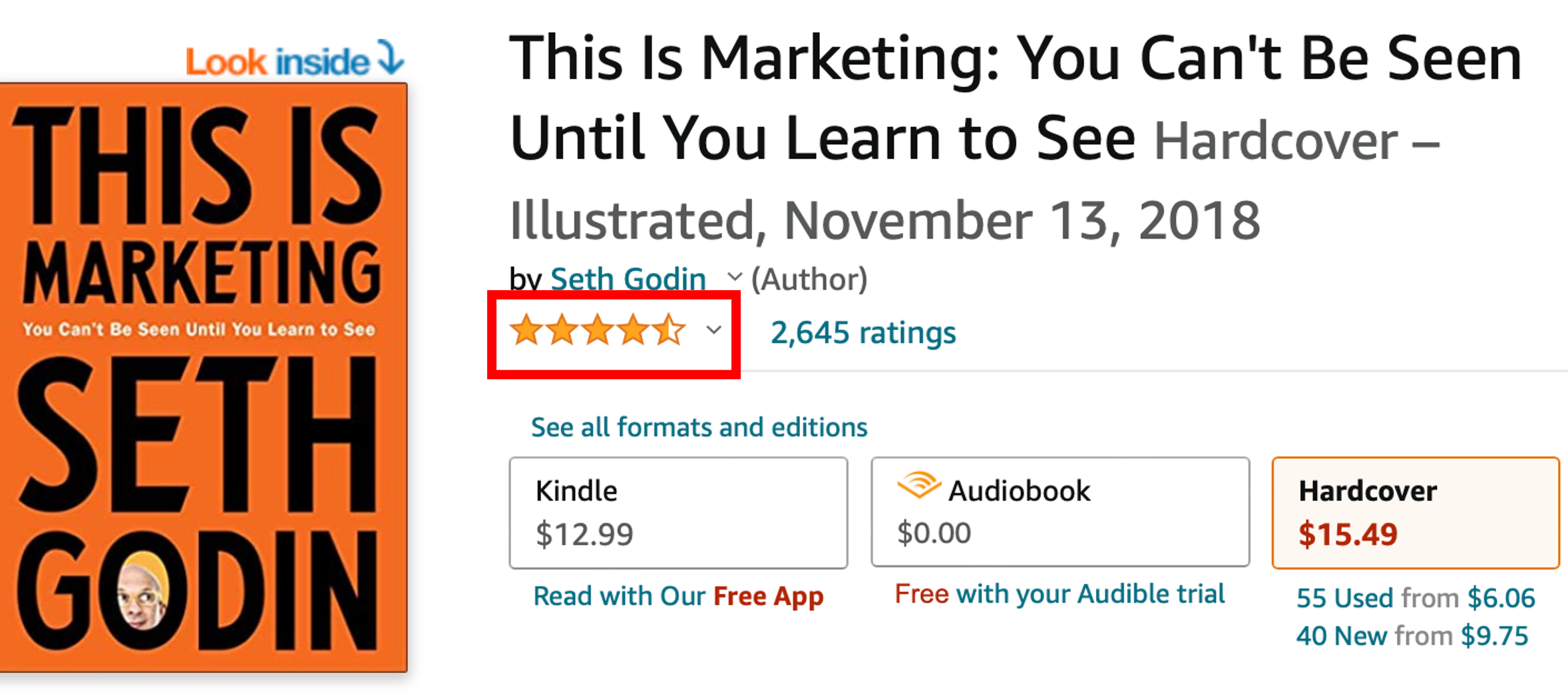}}
\end{minipage} 
\end{tabular}
\end{figure}

In other situations, personalization of rating aggregation is simply not possible. Historical user data may not be available, \eg, in cold-start problems \citep{Padilla.2021}, or data may be sparse, \eg, in settings where customers have only a small number of interactions with a website. For instance, most raters on Yelp provide only a single review (this is true for 71\,\% of the raters in our Yelp dataset), because of which personalization is impeded.  Against this background, there is a vast array of websites that refrain from personalization and instead rely on overall ratings that are intentionally universal across the user base (see \Cref{fig:screenshot}). Examples include movie review platforms (\eg, IMDb, Rotten Tomatoes), comparison websites (\eg, Google Maps, Yelp), booking platforms (\eg, Booking.com), and even shopping platforms (\eg, Amazon, Google Play store).

\begin{figure}[htb]
\OneAndAHalfSpacedXI
\centering
\caption{Example showing that when using the sample mean for rating aggregation, the underlying quality dynamics may not be captured.\label{fig:example_rating_aggregation}}
\includegraphics[width=.5\textwidth]{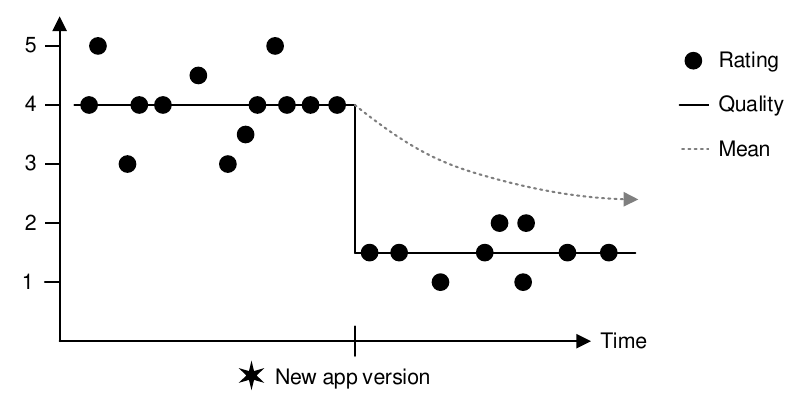}
\end{figure}

In rating aggregation, overall ratings are usually based on the sample mean \citep{Dai.2018,MojicaRuiz.2016,NYT.2019}. However, in this paper, we argue that rating aggregation based on the sample mean is problematic. 
As an illustration, let us take a popular office app that has largely received positive ratings (\eg, 4 out of 5 stars) as an example. Assume that the developer now decides to introduce a subscription model, which customers do not like. As a result, customers rate the \textquote{current} quality of the app with an average of only 2 stars. Due to the many positive star ratings the app received before the introduction of the subscription model, the sample mean adapts very slowly to the new quality assessment (from 4 to 2 stars). Hence, it is evident that the sample mean is a suboptimal indicator of the current quality (see \Cref{fig:example_rating_aggregation}). Another problem is that all ratings are treated the same using the sample mean; there is no differentiation between individual ratings. For instance, review heterogeneity -- some reviews may be more credible -- plays an important role, yet review heterogeneity is not directly considered in the sample mean \citep{Resnick.2000}.

There are various challenges in developing a data-driven model for rating aggregation. (1)~Quality is variable and can change over time \citep{Godes.2012,Li.2008}: app developers publish regular updates and restaurants often change staff, which can have an impact on perceived satisfaction. Hence, \REV{aggregation methods} must be able to reflect the dynamics of ratings over time. (2)~There is substantial heterogeneity among reviews and reviewers. Thus, some ratings may be more informative than others, for example, reviews that are voted as helpful or published by trusted reviewers (\eg, elite status).


\REV{To address the above challenges, we demonstrate the value of using a dynamic, data-driven framework for rating aggregation. For this purpose, we provide an empirical application of a tailored Gaussian process (GP) model, which performs a continuous-time interpolation of ratings over time and is thus naturally dynamic. Furthermore, the GP model considers various covariates that additionally describe the heterogeneity of ratings (\eg, review sentiment, helpfulness, elite status). We also present and compare two estimation strategies, namely, full Markov chain Monte Carlo (MCMC) sampling and scalable variational inference, which offer complementary advantages in terms of theory building and computational efficiency for large-scale online platforms.}


\REV{We evaluate the predictive power of the GP framework in predicting based on 121,123 ratings from Yelp. Specifically, we compare a tailored GP model against the sample mean, a common approach in practice for computing overall rating scores \citep[\eg,][]{Dai.2018,MojicaRuiz.2016,NYT.2019}. We first show that \REV{a custom GP} is superior to the sample mean in terms of predicting future customer satisfaction. We further show that \REV{a custom GP} is superior to common machine learning baselines (\eg, long short-term memory and transformer networks from the realm of deep learning). We also demonstrate the applicability of the GP framework in settings other than restaurant ratings (namely, hotel ratings and app ratings).}


Our research has important implications for marketing practice and research. In marketing practice, it is common to build upon the sample mean for rating aggregation. \REV{However, as argued earlier, the sample mean adapts only slowly to systematic changes in quality, whereas tailored approaches for rating aggregation need to be inherently dynamic. For marketing research, we contribute to the literature on online rating aggregation by empirically demonstrating the value of moving beyond simple means and, for this purpose, show the benefits of dynamic, data-driven aggregation using a custom GP. GPs have only recently been introduced in marketing science \citep[e.g.,][]{Dew.2018,Dew.2020,Dew.2021,Dew.2024,Dew.2024b,Kim.2021} due to their beneficial mathematical properties in modeling customer behavior; however, they were not yet tailored to the setting of modeling online rating sequences. Here, we thus present a novel empirical application aimed at rating aggregation.
}


The remainder of this paper is organized as follows. In \Cref{sec:related_work}, we present related work on rating aggregation. \REV{\Cref{sec:model_development} introduces the GPs framework used for dynamic rating aggregation. In \Cref{sec:setting}, we describe the empirical setting. \Cref{sec:findings} evaluates the predictive performance of the GP framework relative to existing aggregation mechanisms that are commonly used in marketing practice.} In \Cref{sec:discussion}, we discuss our findings and managerial implications, while \Cref{sec:conclusion} concludes.

\section{Related Work}
\label{sec:related_work}

\subsection{Reputation Systems}


Digital marketplaces, where people can exchange goods, services, or information, are one of the great success stories of the Internet \citep{Tadelis.2016}. These marketplaces offer the opportunity for exchanges between strangers (suppliers and customers), who, otherwise, might have never met each other. However, unlike in physical stores, where shoppers can test products or services, online customers cannot test products or services before making a purchase, which causes uncertainty about the quality of the product or service \citep{Resnick.2000}. To mitigate this uncertainty, people are thus highly dependent on online feedback -- often online ratings -- which helps them decide whom to trust \citep{Ba.2002,Cho.2022}. Thus, to foster trust, marketing makes widespread use of reputation systems to inform customers about the quality of goods and services \citep{Tadelis.2016}. This is now considered an essential component of well-functioning online marketplaces and thus how companies exchange information with potential customers \citep{Fradkin.2018}.


A reputation system has been defined as a system that \textquote{collects, distributes, and aggregates feedback about participants' past behavior} \citep{Resnick.2000}. It is common that reputation systems consolidate multiple reports of customer satisfaction into a single overall score \citep{Garcin.2009}, so that customers can compare different entities (\eg, products, restaurants, movies). Hence, in many cases, marketers use such systems to aggregate ratings and present some form of overall star rating. Typically, reputation systems with such overall ratings are \emph{user-independent}. As such, they provide an identical overall rating for all users, which enables the global comparison across products or services. Examples are movie review platforms (\eg, IMDb, Rotten Tomatoes), comparison websites (\eg, Google Maps, Yelp), booking platforms (\eg, Booking.com), and shopping platforms (\eg, Amazon, Google Play store). These websites purposefully avoid personalization to allow for comparisons. In other situations, websites cannot provide personalization for privacy reasons or because past consumption data are lacking, for example, when customers rate a restaurant for the first time (this is true for 71\,\% of the customers in our Yelp dataset) or when the collection of consumption data is intentionally not part of the business model (\eg, IMDb). In this paper, we focus on the setting of aggregating ratings into overall ratings that are user-independent.


Reputation systems provide an overall score, which is in stark contrast to recommender systems. Recommender systems infer individualized preferences from past consumption data so that products or services that have a high likelihood of being purchased can be recommended to customers \citep[\eg,][]{Kahng.2011,Karatzoglou.2010}. Such recommendations are typically used on two locations of a shopping website: (i)~when users search for products, whereby the search is personalized; and (ii)~when recommending related products (\eg, when Amazon displays products that are frequently purchased together), but \emph{not} when computing overall rating scores. Mathematically, recommender systems typically output \emph{rankings}, whereas rating aggregation outputs overall \emph{star ratings}. Hence, recommendation algorithms fulfill a different purpose for marketing professionals and are not directly applicable for aggregating ratings into an overall score. There is another crucial difference: recommender systems require users to ``sign in,'' while most review platforms do not require users to ``sign in,'' so they ``do not know'' the users. Hence, this is another practical reason for using aggregate ratings rather than individually customized ratings.

\subsection{Reputation System Variables: Valence and Review Heterogeneity}

To increase trust and reduce uncertainty in online interactions, reputation systems use various variables (see \Cref{fig:cues_reputation_systems}). These variables describe the valence (\ie, whether a user perceives a service positively or negatively), yet other variables may also capture the heterogeneity among reviews.

\begin{table}[htb]
\centering
\SingleSpacedXI
\footnotesize
\caption{Example of reputation system variables.\label{fig:cues_reputation_systems}}
\begin{tabular}{l | lp{9.5cm} | l}
\toprule
Group & \multicolumn{2}{l}{Examples} & Used in our framework \\
\midrule
{Valence} & $\bullet$ & Numerical rating score \citep[\eg,][]{Archak.2011,Chevalier.2006,Chintagunta.2010,Dellarocas.2007,Forman.2008,Godes.2004,Li.2008,Liu.2006,Moe.2011,Zhu.2010} & Emission \\
\midrule
\lcellt{Review\\ \mbox{heterogeneity}} & $\bullet$ & Review sentiment \citep{Archak.2011,Tirunillai.2014,Netzer.2012} & GP mean \\
 & $\bullet$ & Mean rating score per user \citep{Dai.2018} \\
 & $\bullet$ & Helpfulness \citep{Weiss.2008}  \\
 & $\bullet$ & Review length \citep{Chevalier.2006,Hu.2017} \\
 & $\bullet$ & Temporal contiguity \citep{Chen.2013} \\
& $\bullet$ & Experience such as time on platform or elite status \citep{Chen.2013,Dai.2018} \\
& $\bullet$ & Textual cues such as linguistic modality \citep{Yu.2010} \\ 
\bottomrule
\end{tabular}
\end{table}


At the core of many reputation systems are online ratings. Online ratings (\eg, star ratings) have been extensively studied in previous literature. For instance, the relationship between ratings and sales (rank) was studied \citep[\eg,][]{Archak.2011,Chevalier.2006,Chintagunta.2010,Dellarocas.2007,Forman.2008,Godes.2004,Li.2008,Liu.2006,Moe.2011,Zhu.2010}. Some reviewers leave systematically lower ratings \citep[\ie, stringency; see][]{Dai.2018}; we account for this by using the average rating per user. In addition to the numerical rating score, the review text may express a positive or negative sentiment  \citep{Tirunillai.2014}. As with numerical ratings, the valence of review texts also correlates with sales \citep{Archak.2011}. Our research, however, differs fundamentally from these previous analyses. In most of the abovementioned studies, a certain variant of an aggregated rating score (\eg, average valence or sentiment) serves as \textit{input} (\eg, to study sales); in our work, the (expected) rating score is \textit{output}.


In addition to valence cues, modern reputation systems use several mechanisms to stimulate trust between strangers by displaying further information on reviews or reviewers. This information can be used to model heterogeneity among reviews and reviewers. Review heterogeneity, for example, may encode for whether a user has found a review helpful \citep{Cao.2011,Mudambi.2010,Sipos.2014}. Other heterogeneity sources are the length of the review text, which is a measure of information quality \citep{Chevalier.2006,Hu.2017}, and temporal contiguity, that is, a review that was written shortly after using the product or service \citep{Chen.2013}. In the case of temporal contiguity, a shorter delay between consumption and authoring a review should provide a more exact description of the quality. Reviewer heterogeneity describes sources that relate to the authors of reviews and their credibility. These include, for example, their experience (time on platform or elite status). Experienced reviewers -- elite users -- are said to observe quality signals with higher precision \citep{Chen.2013,Dai.2018}; however, at the same time, their ratings are biased toward past crowds, probably because they care about their social status \citep{Dai.2018}, lowering the credibility of their reviews. Another source of heterogeneity stems from a reviewer's helpfulness \citep{Weiss.2008}. In addition, textual cues, and thus the narrative of the review, may also be informative. In particular, text mining can be used to infer the uncertainty in the perception of the reviewer. For example, \citet{Yu.2010} analyzed words classified as modal auxiliaries such as ``can,'' ``should,'' etc., as this may signal uncertainty. \REV{Although various variables on review heterogeneity can indicate that a rating is particularly important, these variables are typically omitted when aggregating ratings (\ie, all ratings are treated the same when computing the sample mean, although some ratings might be more informative). This weakness is addressed by our proposed framework, in which both the dynamics of online ratings and review heterogeneity variables are explicitly considered.}

\subsection{Rating Aggregation}


To help inform customers about a product's expected quality, reputation systems process long rating sequences into a single aggregated rating score, the overall rating \citep{Dai.2018}. Such an overall rating must fulfill a key requirement: it must be user-independent so that it can be displayed directly on a website to arbitrary visitors. Examples include Yelp, Amazon, and IMDb, which present a universal rating for the entire user base. Many customers rely on these ratings when making purchasing decisions, yet little research has gone into developing approaches that systematically condense rating histories into an overall rating. In practice, the overall rating is often computed by a sample mean (cf. \Cref{tbl:practice}). If a product or service is subject to constant quality, then the sample mean has obvious advantages: it provides an unbiased, i.i.d.  signal of the true quality \citep{Dai.2018}. However, the sample mean has shortcomings when the underlying quality of products or services is dynamic \citep{MojicaRuiz.2016}. This was also confirmed in the study by \citet{Dai.2018}, in which the authors reported an explanatory (but not predictive) model for describing past rating behavior.

\begin{table}[H]
\OneAndAHalfSpacedXI
\centering
\caption{Rating Aggregation in Marketing Practice.}	
\label{tbl:practice}
\small
\begin{tabular}{l l p{6cm} l}
\toprule
\textbf{Platform} & \textbf{Aggregation mechanism} & \textbf{Input variables} & \textbf{Reference} \\
\midrule
{Amazon} & Sample mean$^\dagger$ & Ratings & \citep{NYT.2019} \\
{Google Play} & Weighted mean & Ratings & \citep{Glick.2019} \\
{IMDb} & Weighted mean & Ratings & \citep{IMDb.2025}\\
{Yelp} & Sample Mean$^\ddagger$ & Ratings & \citep{Dai.2018} \\ 
\midrule
\emph{This paper} & GP model & Ratings (with timestamp), heterogeneity variables (e.g., helpfulness) &  \\ 
\bottomrule
\multicolumn{4}{l}{\tiny $^\dagger$ With proprietary machine learning-based post-processing} \\
\multicolumn{4}{l}{\tiny  $^\ddagger$ Subject to a proprietary filtering if not forbidden by a country's legal framework} \\ 
\end{tabular}
\end{table}

In order to improve over the sample mean, alternative approaches for aggregating ratings that put more emphasis on recent ratings have been explored in the literature. Examples include a sliding window mean \citep{Ivanova.2017}, a discounted mean \citep{Garcin.2009,Leberknight.2012}, or a weighted mean \citep{IMDb.2025}. However, in these approaches, both valence and review heterogeneity variables beyond the rating are neglected. \REV{In contrast, we propose to leverage the GP framework for rating aggregation and empirically demonstrate the performance when applying GPs to this task.}

\section{Model Development}
\label{sec:model_development}

\subsection{Problem Statement}


Let $\mathbb{E}[R]$ denote the expected rating. For customers, this provides intuitive information related to the underlying quality of a product or service; thus, it should allow customers to estimate their prospective satisfaction. The expected value $\mathbb{E}[R]$ represents the output of the prediction model. In practice, the output can be a star rating $R \in \{1, \ldots, 5\}$ (as in the case of Yelp) or any other form of numerical feedback (\eg, a relative score $R \in [0, 1]$ of up/down votes, as in the case of YouTube). 


The input should incorporate past online ratings and review heterogeneity variables in a comprehensive manner: it comprises the star rating history for an entity (\ie, product, service, business). Note that the length of the rating history differs across entities; this must be considered in the model specification. The ratings originate from different timestamps (i.e., measured in continuous time), whereby old ratings should a priori be less relevant than recent ratings. The prediction model should also accommodate further variables describing review heterogeneity. 


Based on the above setting, \REV{the} prediction model needs to fulfill three requirements. First, the perceived quality of a rating entity is variable over time. Analogously, a prediction model must capture such dynamics by modeling the actual rating sequence over time. Second, the prediction must consider additional variables concerning both review heterogeneity variables. Third, at deployment time, the expected rating $\mathbb{E}[R]$ must eventually be computed using only data from \emph{past} ratings and \emph{without} access to variables from prospective reviews. \REV{To address these challenges, we adapt the GP framework \citep[e.g.,][]{Dew.2018,Rasmussen.2006} to our empirical application, namely, rating aggregation.}

\subsection{Model Specification}
\label{sec:specification} 

\subsubsection{Overview.}


\REV{This section adapts the GP framework to our empirical application: that is, modeling the continuous dynamics of online ratings over time for the purpose of computing aggregated ratings that predict customer satisfaction.}

\REV{In our empirical application, we use a tailored GP model that has five components:} (1)~an observation sequence given by ratings; (2)~a latent GP; (3)~an emission component linking ratings and the latent GP; (4)~a model for the GP mean function; and (5)~a kernel. The GP is key, as it is responsible for performing a time-dependent interpolation. At the same time, it comprises a mean function that accommodates additional variables. \Cref{fig:schematic_overview,fig:plate_notation} presents a schematic view of the model, which we detail in the following.

\begin{minipage}[t]{.6\textwidth}

\begin{figure}[H]
\centering
\SingleSpacedXI
\caption{Schematic overview of the tailored GP model.}
\label{fig:schematic_overview}
\includegraphics[width=\linewidth]{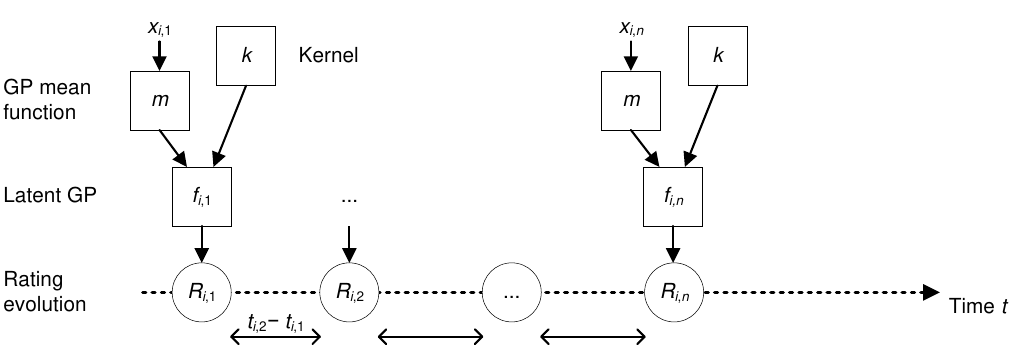}
\footnotesize
\begin{tabular}{p{8cm}} \emph{Note:} The plot shows the latent Gaussian process (GP) model for a given entity $i$ (e.g., a product, service, or company to be rated). The evolution of ratings, which we aim to model, is directly observed. This is achieved by introducing a latent function $f$ in the form of a Gaussian process. The Gaussian process itself is specified by two components, namely, the mean function $m$ and the kernel $k$.
\end{tabular}
\end{figure}

\end{minipage}%
\begin{minipage}[t]{.4\textwidth}
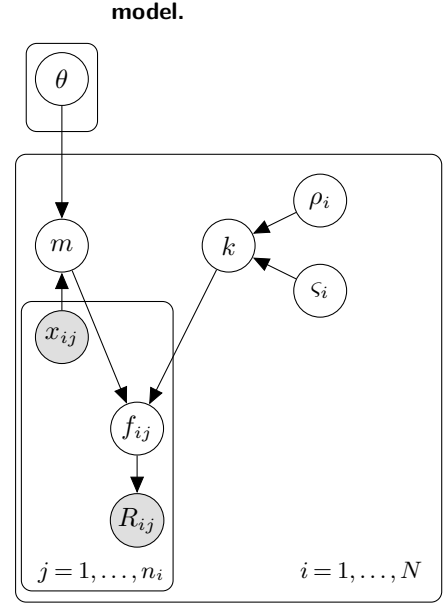
\begin{figure}[H]
	\centering
	\caption{Plate notation of the tailored GP model.} 
	\label{fig:plate_notation}	
	\footnotesize
	\begin{tikzpicture}[latent/.append style={node distance=.5}]
		\node[latent](Theta){$\theta$};
		\node[latent,below=1.5 of Theta](Mean){$m$};
		\node[latent,right=1.5 of Mean](Covariance){$k$};
		\node[latent,right=of Covariance,yshift=0.6 cm](Lengthscale){$\rho_i$};
		\node[latent,right=of Covariance,yshift=-0.6 cm](Variance){$\varsigma_i$};

		\node[obs,below=of Mean] (Valence Cues) {$x_{ij}$};%
		\node[latent,below=of Valence Cues,xshift=1.0cm](Function){$f_{ij}$};
		\node[obs,below=of Function] (Rating) {$R_{ij}$};%

		\plate [] {plate1} {(Rating)(Valence Cues)(Function)} {$j=1,\ldots,n_i$}; %
		{\tikzset{plate caption/.append style={xshift=1cm}}
		\plate [inner sep=0.25cm,minimum width=4cm] {plate2} {(Rating)(Valence Cues)(Variance)(Lengthscale)} {$i=1, \ldots,N$};
		}
		\plate [] {plate3} {(Theta)} {};
		
		\edge {Function}{Rating};
		\edge {Theta}{Mean};
		\edge {Valence Cues}{Mean};
		\edge {Lengthscale}{Covariance};
		\edge {Variance}{Covariance};
		\edge {Mean}{Function};
		\edge {Covariance}{Function};
\end{tikzpicture}		
\end{figure}
\end{minipage}

\subsubsection{Observation Sequence.} 


We model a sequence of ratings for a given entity (e.g., a product, service, or company to be rated). We refer to entities via variable $i$. Each entity comes with a sequence of past ratings that we denote by $R_{i,1}, \ldots, R_{i,n_i}$, where $n_i$ is the number of ratings for entity $i$. The number of ratings per entity $n_i$ is not constant but varies across entities. This has important implications, as it necessitates a flexible model that can handle input sequences of variable size. For instance, only a few ratings might have been collected for a new entity, whereas established entities often have several hundred ratings. 


The ratings $R_{ij}$, $j = 1, \ldots, n_i$, originate from a distribution $\mathcal{R}$. We do not make any explicit assumption on the specific distribution of $\mathcal{R}$; that is, it can be a discrete set (\eg, when handling star ratings) or any other distribution. For instance, for Yelp data, $\mathcal{R} = \{ 1, \ldots, 5 \}$ is used for star ratings from 1 to 5 stars. For ease of notation, we state the model for a discrete set of star ratings in the following. For other representations of customer feedback, a marketer merely needs to update the ordered probit in the emission function to the new distribution. Each rating is further associated with a time $t_{ij}$ and covariates $x_{ij}$ that describe the review heterogeneity (\eg, the sentiment of the written review, helpfulness votes), which enter the GP mean function. 

\subsubsection{Latent Gaussian Process.}


We model the dynamics behind a rating sequence via a latent Gaussian process. It is responsible for performing a dynamic interpolation. Formally, let $f_i$ refer to the latent Gaussian process of entity $i$. 
We provide the rationale for using (1)~latent dynamics and (2)~Gaussian processes in the following.


Modeling the interpolation via latent dynamics is based on prior theory according to which rating represents stochastic signals of quality \citep{Luca.2011}. This is reflected in the model specification, where we directly model ratings as stochastic realizations of the latent GP. 
Including latent dynamics in the prediction model is also beneficial from a practical point of view: latent dynamics are highly effective in modeling quality for which the course is variable \citep{Netzer.2008}. Because of this, predictions based on latent dynamics are often less sensitive to noise. 


Gaussian processes are a common choice for interpolations, as their use comes with obvious mathematical benefits \citep{Dew.2018}: first, interpolations based on Gaussian processes are naturally smooth (i.e., in continuous time) and can thus fit arbitrary curves. Second, Gaussian processes leverage the distance between observations during interpolation. In the context of our work, this allows us to consider the actual temporal differences between ratings. As a result, we can account for a rating that was recorded a year ago vs. current ratings with greater temporal proximity, \eg, a month or a week ago. \REV{Third, Gaussian processes can be further tailored to accommodate additional review heterogeneity variables within the GP mean function. Despite their many advantages, Gaussian processes have only recently received traction in marketing research \citep{Dew.2018,Dew.2020,Dew.2021,Dew.2024,Dew.2024b,Kim.2021}, and, here, we present a novel empirical application, namely, rating aggregation.}


Mathematically, a Gaussian process is a stochastic process that, at each point, follows a normal distribution \citep{Rasmussen.2006}. Hence, it can be seen as a generalization of multivariate normal distributions. Then, a Gaussian process $\mathcal{GP}(m,k)$ is specified by a mean function $m$ and a kernel $k$ \citep{Rasmussen.2006}. There are no specific requirements with regard to the mean function; however, the kernel must be positive semidefinite. Here, we refer to \citet{Dew.2018} for an excellent overview of suitable kernels. 


\REV{In our modeling approach, each entity has its own latent Gaussian process $f_i$ given by $\mathcal{GP}(m,k)$. The mean function $m$ is later formalized as a linear model (see \Cref{sec:GP_mean}), \REV{with shared coefficients $\theta$ across restaurants. We intentionally choose shared coefficients $\theta$ due to several reasons. First, it stabilizes coefficient estimates by leveraging information across all restaurants. Second, it avoids overfitting when individual restaurants have limited data. Third, it yields interpretable coefficients of review-level covariates that are consistent across all restaurants}. As additional notation, we introduce $f_{ij}$, which refers to the value of the latent GP when evaluated at the time of the rating $R_{ij}$.}

\subsubsection{Emission Component Linking Latent GP and Ratings.}

The emission component links the latent GP to the observable rating score. Here, the idea of \textquote{emissions} is similar to that in other latent state models such as hidden Markov models \citep[\eg,][]{Abhishek.2012,Ascarza.2013,Ascarza.2018,Ding.2015,Li.2011,Montgomery.2004,Montoya.2010,Naumzik.2021,Netzer.2008,Schwartz.2014,Schweidel.2011,Zhou.2022}. It specifies the probability of observing a certain rating conditional on the (continuous) value of the latent GP. As such, ratings are stochastically linked to the latent GP.

\REV{We model the} emissions as an ordered probit analogous to earlier works in the rating literature \citep[\eg,][]{Moe.2012}. Based on this choice, the continuous values from the latent GP are mapped onto discrete ratings. Formally, an emission is given by $P\left( R_{ij} = r \mid f_{ij}\right)$, which specifies the conditional probability of observing a specific rating $r \in \{ 1, \ldots, n_r \}$ given the value $f_{ij}$ of the latent GP. Accordingly, $P\left( R_{ij} = r \mid f_{ij}\right)$ is set to an ordered probit. 

\subsubsection{GP Mean Function.} 
\label{sec:GP_mean}

The mean function $m$ specifies the expected value of the latent function $f_i$. It accommodates additional covariates $x_i$ that describe the heterogeneity amongst reviews beyond the numerical rating score. Here, we formalize the mean function as a linear model 
\begin{equation}
m(x_{ij}) = \theta^T x_{ij} 
\end{equation}
with coefficients $\theta$. Hence, if the covariates are positively (negatively) linked, then the GP mean is increased (decreased). \REV{While the temporal GP dynamics remain entity-specific, the coefficients $\theta$ are shared across entities. We intentionally choose shared coefficients for several reasons. First, pooling stabilizes estimation by leveraging information across all restaurants, which is particularly important when the restaurants have a varying number of past reviews. Second, shared coefficients prevent overfitting by avoiding the estimation of separate coefficients for each restaurant. Third, shared coefficients yield interpretable associations between review-level covariates and the mean function across all restaurants. Therefore}, we can later interpret the coefficients $\theta$ to assess which covariates actually affect the mean function (and thus the overall prediction) and in which direction \REV{(e.g., ceteris paribus, a negative coefficient corresponds to a negative association between the corresponding covariate and the aggregated rating)}. \footnote{\REV{The GP model can also be estimated separately for each entity to have entity-specific coefficients $\theta_i$.}}

\REV{In addition to the general review-level variables, the mean function additionally includes the reviewer's mean rating as a covariate to control for stable cross-user heterogeneity in rating tendencies. This term reflects persistent differences in leniency or stringency across reviewers, which prior work has shown to be relatively stable in online review settings \citep[e.g.,][]{Dai.2018}. Importantly, this covariate serves a different purpose than the temporal aggregation problem discussed earlier: while the sample mean of past ratings is problematic for aggregating time-varying entity quality, the reviewer's mean rating is used solely as a bias-correction term for stable reviewer-specific tendencies.}

\subsubsection{Kernel.}
\label{sec:GP_covariance}


The kernel in the GP model accommodates the timing of the ratings. Following the recommendations in \citet{Dew.2018}, we decided to use an exponential kernel function for modeling the covariance function.\footnote{\REV{We also considered a wide range of other covariance functions in Online Appendix~\ref{appendix:robustness_checks}, including a squared exponential (SE) kernel and a rational quadratic kernel.}} The exponential kernel function is a special case of the Matérn class of kernel functions \citep{Rasmussen.2006}. It is fairly flexible, allows for a certain roughness in the GP, and should thus be ideally suited in the context of discrete ratings. Accordingly, the covariance between the two realizations of $f_i$, \ie, $f_{ij}$ and $f_{ik}$, is specified as
\begin{equation}
\label{eq:non_stationary_kernel}
COV(f_{ij}, f_{ik}) = k\left(t_{ij}, t_{ik}\right) = \sigma_i^2 \, \exp\left( -\frac{1}{\rho_i} \abs{t_{ij}-t_{ik}} \right), 
\end{equation}
with entity-specific parameters $\rho_i$ and $\sigma_i$.\footnote{\SingleSpacedXI\footnotesize We emphasize that the covariance function is indeed positive semidefinite and thus a valid kernel; see \citet{Rasmussen.2006}.} \REV{The parameter $\rho_i$ is called the \emph{characteristic length scale}, and the hyperparameter $\sigma_i > 0$ is called the \emph{amplitude} \citep{Rasmussen.2006}.} In other words, $\rho_i$ models how quickly the Gaussian process varies. \REV{Larger characteristic length scales produce slowly varying functions, while shorter length scales produce more \emph{wiggly}, that is, rapidly varying functions. Hence, the larger the $\rho_i$, the more persistent the temporal correlation between ratings, as correlations decay more slowly over time.} In contrast, the variance $\sigma_i^2$ determines the average distance of the function away from its mean.


The kernel in \Cref{eq:non_stationary_kernel} accommodates the time difference between ratings (\ie, $\abs{t_{ij}-t_{ik}}$). This term is associated with a negative sign so that ratings in close temporal proximity are allowed to have a larger covariance compared to distant ratings. Hence, all else being equal, longer time intervals between ratings lead to a small covariance. If $\rho_i$ is small, the influence of the time difference is large; thus, it can fluctuate more strongly and lead to more local variability and larger ``jumps.'' If $\rho_i$ is large, then the influence of the time difference is small, thus leading to more stable, slowly varying functions. The term $\sigma_i^2$ captures the variance around the mean function and is thus also called signal standard deviation. If $\sigma \rightarrow 0$, the GP will largely mirror its mean function, implying that, when brought to data, our model would be able to perfectly predict customer satisfaction from the data (i.e., review heterogeneity variables).

\subsection{Combined Model}


By combining the above components, we obtain the following latent GP model:%
\begin{subequations}
\begin{align}
m(x_{ij}) &= \theta^T  x_{ij}, && \text{for all } i,j \\
k\left(t_{ij}, t_{ik}\right) &= \sigma_i^2 \, \exp\big( -\frac{1}{\rho_i} \abs{t_{ij}-t_{ik}} \big) , && \text{for all } i,j,k \\
f_i & \sim \mathcal{GP}(m, k), && \text{for all } i \\
R_{ij} \mid f_{ij} & \sim \text{OrderedProbit}, && \text{for all } i,j 
\end{align} 
\end{subequations}
where \textquote{$\sim$} denotes variables that follow a given prior. \REV{The} GP model has several parameters that need to be estimated from data, namely, $\theta$ inside the mean function and $\rho_i$ and $\sigma_i$ inside the covariance function. We provide the so-called plate notation in \Cref{fig:plate_notation}. \REV{The parameters can be estimated using different strategies, depending on whether the purpose is on uncertainty quantification to interpret (latent) parameters for marketing insights or the purpose is on computational scalability to enable deployment at large platforms in practice. In the following, we thus outline two complementary approaches.}

\REV{
\subsection{Estimation Approaches}
\label{sec:estimation_strategies}

We employ two complementary estimation strategies that both estimate the \emph{same} latent GP model from above: (1)~full Bayesian estimation via Markov chain Monte Carlo (MCMC) and (2)~scalable variational inference (VI). Both approaches estimate the same underlying model but differ in computational efficiency and purpose, as follows:

\begin{enumerate}
\item \textbf{MCMC.} The so-called full MCMC approach \citep[see][]{Gelman.2014b} estimates the \emph{full} posterior distribution and thus offers rigorous uncertainty quantification for all parameters, including those \emph{within} the mean function and the GP kernel. This estimation strategy is particularly valuable for generating new insights for marketing. However, full MCMC scales with $\mathcal{O}(n_i^3)$ and is thus not tractable for large datasets. We report the full specification of priors, hyperparameters, estimation details, posterior summaries, and parameter recovery tests in the Online Appendix~\ref{appendix:estimation_details}. The code is provided in Online Appendix~\ref{appendix:source_code}.
\item \textbf{Scalable VI.} The Scalable VI approach \citep[e.g.,][]{Hensman.2015} offers a computationally efficient alternative to MCMC that approximates the posterior distribution, yet at the cost of sacrificing the uncertainty quantification for parameters by only estimating point estimates. Hence, the VI approaches dramatically reduce computation time and can therefore be applied to large-scale datasets, such as those of online platforms with millions of reviews. Because of this, the VI approach is especially relevant in real-world applications where we expect our tailored GP model to be updated after regular time intervals (e.g., monthly).
\end{enumerate}

Our VI approach draws upon ideas from \citet{Hensman.2015} and relies on two simplifying assumptions. First, the posterior distribution $p(f_i \,\mid\, Y_i)$ of the latent function $f_i$, given the observed rating sequence, is modeled as a multivariate Gaussian distribution $q(f_i)$, the parameters of which are trained by maximizing a lower bound of the log-likelihood. Gaussian distributions are widely used in machine learning such as in, \eg, linear models \citep{Hastie.2013}. Second, only a subset of the full rating history $M\subset\{1,\ldots,n_i\}$ is considered; the latter yields a so-called sparse Gaussian process and is a common assumption for estimating GPs in the literature \citep{Hensman.2015}. 

Let us denote the realization of the latent function $f_i$ at $M$ by $u_i$. We further define $q(f_i) = \int p(f_i \,\mid\, u_i) \, q(u_i) \dd u_i$. Then the following theorem gives an evidence lower bound (ELBO) for the likelihood of the GP model.
\begin{theorem} 
For our tailored GP model, the evidence lower bound (ELBO) of its likelihood is given by
\begin{equation}
\label{eq:elbo}
\log p(Y_i)\geq\mathbb{E}_{q(f_i)}\lbrack\log p(Y_i \,\mid\, f_i)\rbrack - \mathrm{KL} \lbrack q(u_i) \,\mid\mid\, p(u_i) \rbrack .
\end{equation}
\end{theorem}

\textsc{Proof.} Follows from \citet{Hensman.2015}. A step-by-step derivation is in Section~\ref{appendix:proof}.\hfill\Halmos

\noindent
For a multivariate Gaussian approximation $q(u_i)$, the Kullback-Leibler divergence in \Cref{eq:elbo} can be calculated explicitly. The first component $\mathbb{E}_{q(f_i)}\lbrack\log p(Y_i \,\mid\, f_i)\rbrack$ can be approximated via numerical integration \citep[see][for details]{Hensman.2015}. This thus allows for efficient estimation, as shown in the following proposition. 

\begin{proposition}
Compared to the full MCMC approach, the computational complexity of our scalable variational inference is reduced from $\mathcal{O}(n_i^3)$ down to $\mathcal{O}(n_i \, \abs{M}^2)$.
\end{proposition}

\noindent
Given a properly chosen $M$ with $\abs M \ll n_i$, this provides a scalable computation scheme for large rating datasets. In our experiments, we found that $M=250$ yielded satisfactory results.

}

\subsection{Differences from Other Modeling Approaches}

We use a latent GP (not to be mistaken with a latent Gaussian normal). GPs have been recently introduced in the marketing literature \citep{Dew.2018,Dew.2020,Dew.2021,Dew.2024,Dew.2024b,Kim.2021}. \REV{In our work, we incorporate additional covariates within our GP mean function}. Nevertheless, we tested a na{\"i}ve GP without the inclusion of such covariates. \REV{Here, no parameters are shared across restaurants, and, hence, the na{\"i}ve GP is fully unpooled, representing an entity-by-entity estimation. We found its performance to be inferior.}


The GP specification in our work is markedly different from the growing body of latent state models in management \citep[\eg,][]{Abhishek.2012,Ascarza.2013,Ascarza.2018,Ding.2015,Li.2011,Montgomery.2004,Montoya.2010,Naumzik.2021,Netzer.2008,Schwartz.2014,Schweidel.2011,Zhou.2022}. In this stream of literature, hidden Markov models, where the latent space consists of a discrete number of states, have been developed. In contrast, our GP model involves a continuous latent state, which is necessary to make accurate inferences over continuous values $\mathbb{E}[R]$. One might think that a hidden Markov model could be theoretically applied to our task. However, a discrete latent state space cannot fully represent predictions $\mathbb{E}[R]$ over a continuous interval. One could theoretically apply a discretization to $\mathcal{R}$ in a hidden Markov model; however, this is mathematically infeasible: for $n_r = 5$ stars and a discretization into 0.1-step buckets, the state space would explode to 50 dimensions with more than $50 \times 50$ transitions, exceeding the available data in most applications. Accordingly, our research objective requires latent dynamics over continuous values, the latter of which is provided by a GP.


\REV{Beyond the literature on latent state models, our tailored GP model is also different from Brownian motion as another continuous-time stochastic process that has been widely applied in marketing, economics, and finance, namely, Brownian motion \cite[\eg,][]{Branco.2016}. While both are continuous-time, continuous-state processes with Gaussian properties, Brownian motion represents a special case of a GP with zero mean and a covariance function of the form $\min(t_i, t_j)$, resulting in a non-stationary kernel and independent increments. In contrast, our tailored GP employs an exponential kernel with entity-specific lengthscale $\rho$, which induces smooth, mean-reverting temporal correlations rather than pure random walk behavior. Moreover, we condition the GP on observed covariates to infer latent trajectories that best explain heterogeneity in the ratings, which goes beyond simply simulating paths. Importantly, Brownian motion is completely defined by time and has no mechanism to include covariates. Finally, our GP uses an ordered probit to model rating emissions.}


For mathematical reasons, other model specifications are also precluded. \REV{For instance, state-space models provide latent dynamics over continuous values and can, in principle, be projected over irregular time intervals. However, incorporating covariates and handling non-equidistant sampling in such models is considerably more complex. In contrast, leveraging the GP framework offers continuous-time interpolation and thus naturally accommodates irregularly spaced observations while maintaining flexibility in including covariates.}

\subsection{Predicting the Expected Rating $\mathbb{E}[R]$}


We now detail how predictions are made using the above GP model. We remind the reader of our particular setting: predictions should potentially be displayed on reputation systems as an aggregated rating score, and, hence, predictions should be \emph{customer-independent} \citep{Dai.2018,Garcin.2009,McGlohon.2010}. For this reason, predictions must \emph{not} be personalized to a specific customer; instead, the outcome must provide a universal score for all -- even unknown -- website visitors. For the same reason, information on valence and review heterogeneity variables of future ratings is unknown at deployment time and cannot be considered for making predictions. Instead, only information from the observed past rating sequence can be used.

\vspace{0.3cm}
\begin{quote}\begin{quote}
\OneAndAHalfSpacedXI
\textbf{Task:} {Let us first define the rating history $\mathcal{Y} = \lbrace Y_{ij}\rbrace_{j=1}^{n_i}$ for entity $i$ with $Y_{ij} = (R_{ij}, t_{ij}, x_{ij})$. It comprises not only the rating score but also the other covariates that are associated with a review. Then, the objective is to model the expected quality $R$ at some point $\tau$ in the future but \textbf{without} access to the review heterogeneity variables. Hence, the objective is to estimate the expected quality $\mathbb{E}\left[ R \mid \mathcal{Y} \right]$ for some $R$ at deployment using only historic values $\mathcal{Y}$.}
\end{quote}\end{quote}
\vspace{0.3cm}

\noindent
At deployment time, we have access to the past $n_i$ ratings for an entity to make predictions. For this, the prediction is computed by applying the following simple inference rule based on which the expected rating for an entity $i$ is computed. When using this inference, values of review heterogeneity variables for future ratings are \emph{not} considered nor is the timing of the next rating considered. Instead, we \emph{marginalize} over their distribution and thus capture the average review heterogeneity variables for a given entity. This is derived in the following section.

\subsection{Derivation of the Inference Algorithm for Prediction}

To make predictions that represent the aggregated rating, let us first define a tuple $Y_{ij} = (R_{ij}, t_{ij}, x_{ij})$. It comprises not only the star rating but also the other covariates that are associated with a review. 

The objective is to estimate the expected quality $\mathbb{E}\left[ R \mid \mathcal{Y} \right]$ given the rating history $\mathcal{Y} = \lbrace Y_{ij}\rbrace_{j=1}^{n_i}$ for entity $i$. Note that the conditional expected rating $R$ at some point in time $\tau$ in the future is to be modeled. This is achieved by the following two steps. (1)~We first derive the predictive posterior distribution of $R$ conditional on the rating history. The predictive posterior distribution for the new star rating $R$ depends on several parameters, namely, a time $\tau^\ast$ and review heterogeneity variables $x^\ast$. However, these variables are unknown at the time of inference, and, hence, (2)~we subsequently marginalize over these variables to determine $\mathbb{E}\left[ R \mid \mathcal{Y} \right]$.

\subsubsection{Predictive Posterior Distribution.}

We derive the predictive posterior distribution $P\left(R=r\mid \mathcal{Y} \right)$ for a rating $R$ that is hypothetically associated with a given timing $\tau^\ast$ and review heterogeneity variables $x^\ast$. For simplicity of notation, we denote them by $\xi^\ast = (\tau^\ast, x^\ast )$. Then, we want to derive the predictive posterior distribution of $R$ given $\mathcal{Y}$ and $\xi^\ast$.

We begin by deriving the posterior distribution of $f_i$ at $\xi^\ast$, which we denote by $f^{\ast}$, via two steps: (1)~Let $\mathcal{F} = ( f_{i,1}, \ldots, f_{i,n_i})$ be the vector where the latent GP is evaluated at the different ratings. Since $f_i$ is modeled as a Gaussian process, the joint prior distribution of both $f^{\ast}$ and $\mathcal{F}$ is also Gaussian \citep{Rasmussen.2006}. It can thus later be replaced by a Gaussian with some mean $\mu$ and some variance $\nu^2$. (2)~We approximate the posterior distribution $P\left( \mathcal{F} \mid \mathcal{Y} \right)$ by a Markov chain Monte Carlo approach (see Online Appendix~\ref{appendix:estimation_details}). That is, we draw $S$ samples $\mathcal{F}_1,\ldots, \mathcal{F}_S$ from the distribution. 

Combining the previous two steps yields an approximation of the posterior distribution of $f_i^\ast$, which is given as
\begin{align}
\label{eq:posterior_approx}
P\left(f^{\ast}\mid \mathcal{Y} \right)  = & \int P\left(f^{\ast}\mid \mathcal{F} \right)P\left( \mathcal{F} \mid \mathcal{Y} \right) \dif \mathcal{F} \\
\approx & \frac{1}{S}\sum_{s=1}^{S}P\left(f^{\ast}\mid \mathcal{F}_s\right) \\
= & \frac{1}{S}\sum_{s=1}^{S}\phi\left(f^{\ast}\mid\mu_{s},\nu_{s}^2\right),
\end{align}
where $\phi\left(\cdot\mid\mu_{s},\nu_{s}^2\right)$ denotes the Gaussian density function with mean $\mu_{s}$ and variance $\nu_{s}^2$. We note that both $\mu_{s}$ and $\nu_{s}^2$ depend on the samples $\mathcal{F}_s$ and $\xi^\ast$. 

Given the above, the predictive distribution over an ordinal rating $R$ given $\xi^\ast$ and rating history $\mathcal{Y}$ can be approximated as
\begin{align}
\label{eq:posterior}
P\left(R=r\mid\xi^\ast,\mathcal{Y}\right)
=&\int P\left(R=r\mid f^{\ast}\right) P\left(f^{\ast}\mid \mathcal{Y} \right) \;\dif f^{\ast} \\
\approx & \frac{1}{S}\sum_{s=1}^{S}\int P\left(R=r\mid f^{\ast}\right)\phi\left(f^{\ast}\mid\mu_{s},\nu_{s}^2\right) \;\dif f^{\ast}\\
= & \frac{1}{S}\sum_{s=1}^{S}\int \Phi\left(\frac{c_{r} - f^{\ast}}{\kappa}\right) - \Phi\left(\frac{c_{r-1} - f^{\ast}}{\kappa}\right) \, \phi\left(f^{\ast}\mid\mu_{s},\nu_{s}^2\right) \;\dif f^{\ast}\\
=& \frac{1}{S}\sum_{s=1}^{S}\Phi\biggl(\frac{c_{r}-\mu_{s}}{\sqrt{\kappa^2+\nu_{s}^2}}\biggr) - \Phi\biggl(\frac{c_{r-1}-\mu_{s}}{\sqrt{\kappa^2+\nu_{s}^2}}\biggr) ,
\end{align}
where $\Phi$ denotes the cumulative distribution function of the standard normal distribution, $c_{r}$ refers to the cutoff values from the ordered probit, and $\kappa$ refers to its standard deviation \citep{Moe.2012}. Here, the last equality was derived with the help of \citet[Chapter\,3.9]{Rasmussen.2006}. 

\subsubsection{Marginalizing over the Posterior Predictive Distribution.}

Based on the above predictive posterior distribution $P(R=r\mid\xi^\ast,\mathcal{Y})$, we now derive the conditional expectation $\mathbb{E}\left[R\mid \mathcal{Y} \right]$. We begin by observing that
\begin{equation}
\label{eq:marginalization}
P\left(R=r\mid \mathcal{Y} \right) = \int P\left(R=r\mid\xi^\ast,\mathcal{Y}\right) P(\xi^\ast)\dif\xi^\ast.
\end{equation}
However, modeling the multivariate distribution of $\xi^\ast$ with potentially mixed margins is challenging in practice. To address this, we approximate the above integral by sampling $L$ times from the observed ratings features and temporal differences $\delta_l$ between consecutive ratings,\footnote{\SingleSpacedXI\footnotesize \REV{We additionally experimented with applying recency-weighted sampling, which places higher weights on more recent observations. However, this adjustment did not lead to meaningful improvements.}} \ie,
\begin{align}
\label{eq:expectation}
P\left(R=r\mid \mathcal{Y} \right) 
=&\int P\left(R=r\mid\xi^\ast, \mathcal{Y} \right) P(\xi^\ast) \;\dif\xi^\ast\\
\approx & \frac{1}{L}\sum_{l=1}^{L}P\left(R=r \,\middle|\, t_{i,n_i}+\delta_l, x_l, \mathcal{Y} \right) . 
\end{align}
\noindent
Finally, we make inferences as follows: we compute the posterior predictive distribution from \Cref{eq:posterior} and then marginalize over it in \Cref{eq:expectation}. This yields the expected quality $\mathbb{E}[R]$ and thus a prediction of expected customer satisfaction that is aggregated into a single score.

\section{Setting}
\label{sec:setting}

\subsection{Data Description}


Our empirical application is based on online ratings for restaurants. On the one hand, restaurants provide services for which quality varies considerably between restaurants and over time. For instance, service personnel and chefs change often. Hence, inferences from ratings must be based on a dynamic model. On the other hand, service quality (and likewise ratings) is of great interest to customers \citep{Tripadvisor.2017}.  


Our dataset consists of ratings from Yelp in Phoenix, Arizona. Yelp provides a reputation system that allows users to write reviews, but it also collects additional information on the reputation of reviews (\eg, reviews can be voted as \textquote{helpful}) and reviewers (\eg, it awards an \textquote{elite status}). Yelp displays aggregated ratings on both their webpage and in their app. The underlying aggregation mechanism is currently based on the sample mean \citep{Dai.2018}.\footnote{\SingleSpacedXI\footnotesize Yelp applies a filtering step, where all ratings that are believed to be nonauthentic and not trustworthy are removed. Accordingly, our data do not include ratings that Yelp has filtered out.} As detailed earlier, inferences from the sample mean are subject to shortcomings, namely, not being dynamic and not considering the heterogeneity among reviews. Hence, this provides a compelling setting in which we can compare the predictive power of different rating aggregation methods.

\subsection{Variable Description}


Our dataset spans from January 2010 to December 2017 and includes 121,123 ratings from 625 restaurants. The rating sequence of each restaurant is then modeled using the above GP model. The distribution of star ratings is J-shaped, with an average of 3.95 (SD = 1.29) that is skewed to 5 stars. The particular J-shape of the rating distribution was previously noted \citep{Hu.2017}. This observation is reflected in our model specification, where the emission probability is formalized by an ordered probit model \citep[\eg,][]{Moe.2012}. Overall, the star ratings reveal considerable variation across both restaurants and time. All variables are collected at the rating level. By choosing such a granular unit of analysis, we consider dynamics in the covariates over time. 


Beyond online ratings, our tailored GP model is parameterized with a variety of variables that are common in contemporary reputation systems; see \Cref{tbl:model_variables}. \REV{In line with our focus on the temporal dynamics of restaurant ratings, we capture review-specific tendencies via covariates in the GP mean function. Most importantly, we include the mean rating for each user associated to the specific review to capture across-user heterogeneity in rating styles, which behavioral work suggests are much more pronounced than within-user heterogeneity over time \citep{Kane.1995,Weijters.2010}. Specifically, within-user heterogeneity (e.g., a user becoming slightly more critical over time) are typically small and tend to average out across multiple reviews. In our framework, such residual within-rater variation is therefore handled implicitly by the GP kernel, which smooths the latent rating process over time.}\footnote{\SingleSpacedXI\footnotesize \REV{We also explored defining the user mean over short rolling windows to make it more local in time, but this considerably increases estimation noise because most reviewers contribute only a few ratings in short horizons, leading to an unfavorable bias–variance tradeoff.}} Furthermore, several control variables are extracted via text mining. First, we extract the length of the written review, as a longer description may be more informative and thus more precise. Second, we measure the language sentiment in the written reviews \citep{Archak.2011} using the approach from \citet{Martin.2014}. We also measure the linguistic modality via text mining \citep{Yu.2010,Feuerriegel.2025}. Linguistic modality captures words classified as modal auxiliaries in the review text (\eg, ``can'', ``should''), which may signal uncertainty -- or vagueness -- in the perception of the reviewer. Temporal contiguity is determined as in \citet{Chen.2013}. That is, we scan for phrases (\eg, \textquote{today}, \textquote{just got back}) that indicate that the review was written on the day of consumption. \REV{To ensure the reliability of linguistic measures, we follow standard practice and use text mining approaches that are well-defined even for short reviews \citep{Yu.2010,Martin.2014}.}\footnote{\SingleSpacedXI\footnotesize \REV{In our dataset, all but one of the 121{,}123 ratings were accompanied by a written review. For the single case without text, we encode review length as 0 (minimum), sentiment as 0 (neutral), and linguistic modality as 0 (minimum).}} 

Needless to say, our model is not limited to the current selection of variables. If other variables appear relevant in practice, they can simply be incorporated into our model in a straightforward manner. \REV{In Online Appendix~\ref{appendix:robustness_checks}, we further experiment with alternative model parameterizations (e.g., by including the number of prior reviews per reviewer) and with other review textual features (e.g., term frequencies \citep[\eg,][]{Li.2011b,Ling.2014,Wang.2016}).}

\begin{table}[htb]
  \SingleSpacedXI
	\centering
	\sisetup{parse-numbers=false}
	\caption{Model Variables.\label{tbl:model_variables}}
	{\footnotesize
	\begin{tabular}{l p{9.5cm} SS }
		\toprule
		\multicolumn{1}{l}{Variable} & Description  & {Mean} & {SD} \\
		\midrule
		\multicolumn{4}{c}{\textsc{Emissions: Online ratings ($=$ dependent variable)}} \\
		\midrule
		Rating score $R_{ij}$ & Discrete star rating between 1--5 stars that denotes the overall customer satisfaction with the restaurant as perceived by the user  & 3.95 & 1.29 \\
		\midrule
		\multicolumn{4}{c}{\textsc{Covariates: Review heterogeneity variables $x$ ($=$ GP mean function)}} \\	
		\midrule
		Review sentiment & Language sentiment encoding the perceived positivity of the written review as in \citet{Martin.2014} &  0.10 & 0.34 \\
		Mean star rating per user & Mean rating score of filing user across all cities (not just Phoenix), whereby we control whether certain users follow a different rating behavior than others (\eg, by being more negative)  & 3.71 & 0.98 \\
		Helpfulness & Number of votes reflecting how often a user was rated as \textquote{helpful} & 0.85 & 6.17 \\
		Review length & Number of words as a proxy for the depth of the review & 100.82 & 95.87 \\ 
		Temporal contiguity & A binary variable reflecting whether the review was submitted directly after the restaurant visit ($=1$) or with a delay ($=0$), as in \citet{Chen.2013} & 0.07 & 0.26 \\		
		Time on Yelp & Measures the period of the user's experience with the rating platform (in years) & 2.81 & 2.19 \\
		Elite status & Official batch assigned by Yelp to particularly prolific users with \textquote{well-written reviews, high-quality photos, and a detailed personal profile}  & 0.19 & 0.39 \\
		 Linguistic modality & Number of modal auxiliaries and qualifiers analogous to \citet{Yu.2010}  & 0.01 & 0.03 \\		
		\bottomrule
	 \multicolumn{4}{l}{Unit of analysis: rating level (i.e., 121,123 observations)} \\
   \multicolumn{4}{l}{\emph{Notes:} The following count variables enter the model as log: time on Yelp, review helpfulness, review length. }
	\end{tabular}
	}
	\footnotesize
\end{table}

\subsection{Baselines}
\label{sec:baselines}

\subsubsection{Arithmetic Baselines.}

We compare our tailored GP model against a wide range of mechanisms with the purpose of aggregating rating sequences into an overall score.
\begin{quote}\begin{quote}
\begin{enumerate}
\item The \emph{sample mean} has -- despite its simplicity -- substantial predictive power \citep{McGlohon.2010}. It is widespread in practice and is employed by reputation systems at Amazon, Google Play, and Yelp (see \Cref{tbl:practice}). 
\item The \emph{weighted mean} accounts for the different frequencies of star ratings during aggregation \citep{IMDb.2025}. This essentially gives a form of Bayesian averaging. 
\item The \emph{sliding window mean} is the mean over the last $l_i$ ratings of each sequence \citep{Ivanova.2017}. 
\item The \emph{discounted mean} places a weight on each rating so that older ratings are discounted \citep{Leberknight.2012}. 
\end{enumerate}
\end{quote}\end{quote}

\subsubsection{Machine Learning Baselines.}

The above mechanisms entail only a few degrees of freedom (\eg, additional covariates cannot be incorporated). To address this limitation, we draw upon machine learning baselines as follows:
\begin{quote}\begin{quote}
\begin{enumerate}
\setcounter{enumi}{4}
\item A \emph{linear model (feature engineering)} is used along with feature engineering (\ie, the average rating, the standard deviation of the rating sequence, the last 3 ratings, the median rating, the mean absolute deviation of the rating sequence, and the linear trend of the rating sequence). Here, feature engineering helps in mapping the variable-size rating sequence onto a fixed-size, low-dimensional vector representation.
\item A \emph{linear model (covariates)} is used with the last rating and the corresponding covariates from \Cref{tbl:model_variables}. As such, the choice of covariates is identical to that in our proposed model. Additionally, a sample mean and a random effect for users, which implicitly accounts for the fact that different user types may yield different star ratings that are more or less trustworthy, is included.
\item A \emph{random forest (feature engineering)} is used along with the above feature engineering. We use a random forest due to its flexibility and strong out-of-the-box performance \citep{Hastie.2013}.
\item A \emph{random forest (covariates)} is used along with the last rating and the corresponding covariates from \Cref{tbl:model_variables}. As such, the choice of covariates is identical to that in our proposed model. Additionally, a sample mean is again included.
\item A \emph{univariate LSTM} (long short-term memory) that was fed with only the rating sequences and an LSTM with multivariate input (\ie, star ratings, their time). We use an LSTM as a state-of-the-art recurrent neural network for sequence learning \citep{Hochreiter.1997}.  
\item A \emph{multivariate LSTM} is fed with ratings and all other covariates related to valence and review heterogeneity variables. Thus, this LSTM variant has access to the same data as our proposed model.
\item A \emph{transformer} is a specific neural network that weighs different parts of the input sequence through a self-attention mechanism \citep{Vaswani.2017}. Due to their particularly deep network architecture, transformers are currently regarded as the state of the art for modeling complex, high-dimensional sequences \citep[\eg,][]{Liu.2020,Wolf.2020}. For the transformer, we use the same input as in the multivariate LSTM.
\end{enumerate}
\end{quote}\end{quote}
\noindent
Several of the above baselines entail further \mbox{(hyper-)}parameters that must be tuned; these are detailed in Online Appendix~\ref{appendix:baselines_implementation}. \REV{In Online Appendix~\ref{appendix:baselines_cs}, we also compare our tailored GP model against a range of other baselines from computer science, namely (1)~recommender algorithms, (2)~information aggregation (``crowdsourcing''), and (3)~truth discovery. These comparisons confirm that the GP specification from above performs best.}

\section{Empirical Findings}
\label{sec:findings}

\subsection{Estimation Results}

\REV{\Cref{tbl:model_fits} reports the estimation results using the full MCMC estimation strategy from  above. As a result, this estimation strategy allows us to interpret the (latent) coefficients inside the GP, such as how variables from review heterogeneity are linked to the mean function.}

\begin{table}[h!]
\OneAndAHalfSpacedXI
\centering
\def\sym#1{\ifmmode^{#1}\else\(^{#1}\)\fi}
\sisetup{parse-numbers=false,input-symbols={()},table-space-text-post = \sym{***}}
\caption{Estimation Results.}\label{tbl:model_fits}
{\footnotesize
\begin{tabular}{l S S S}
\toprule
Variable & {(1)} & {(2)} & {(3)}  \\			
\midrule

\csname @@input\endcsname lgpm_coefficients

\midrule
\csname @@input\endcsname lgpm_information_criteria
\bottomrule
\multicolumn{4}{l}{Stated: coefficients (std. errors in parenthesis); significance stars: $^{*}$ 0.05, $^{**}$ 0.01, $^{***}$ 0.001}
\end{tabular}
}
\footnotesize
\begin{tabular}{p{12.5cm}} \emph{Note:}  The model fit is compared based on the widely applicable information criterion~(WAIC). The WAIC is reported on the deviance scale, such that smaller values indicate a better model fit. This follows suggestions for Bayesian modeling according to which traditional metrics, such as the deviance information criterion~(DIC) or the Akaike information criterion~(AIC), should be avoided \citep{Gelman.2014}.  
\end{tabular}
\end{table}

In terms of review heterogeneity, several variables are relevant. In line with our expectations, the coefficient belonging to the review sentiment is both positive and statistically significant (coefficient of $0.801$ in column~(3); $p$-value $< 0.001$). Hence, a review with a positive (or negative) tone may contribute further signals that are informative. The mean star rating score of the user is also positively linked and statistically significant (coefficient of $0.275$ in column~(3); $p$-value $< 0.001$). On this basis, the GP model can account for past ratings from users who tend to lean toward negative or positive valence. 
Review length is negatively linked, which could be attributed to dissatisfied users who may tend to write more lengthy reviews to verbalize their complaints. Temporal continuity refers to reviews that are written shortly after consumption. While one could expect that this would correspond to ratings that are more predictive, we find a negative and statistically significant coefficient. Reviewer experience -- as measured by the users' amount of time on Yelp -- is associated in a positive direction (coefficient of $0.109$ in column~(3); $p$-value $< 0.001$). Put simply, it appears that a customer with a larger tenure is apt to give ratings that are better in predicting the current quality. In addition, ratings from a user with elite status are associated negatively with the predicted customer satisfaction (coefficient of $-0.268$; $p$-value $< 0.001$).

\Cref{tbl:model_fits} further reports the model fit (row: WAIC). The results help us assess how relevant additional variables are for describing the evolution of ratings. We adhere to the latest recommendations in Bayesian modeling \citep{Gelman.2014}, according to which traditional metrics, such as the deviance information criterion~(DIC) or the Akaike information criterion~(AIC), should be avoided. Instead, preference should be given to metrics that are better suited for mixture and hierarchical models. Hence, we report the widely applicable information criterion~(WAIC). In the na{\"i}ve GP from column~(1), all covariates are absent; however, it can still draw upon the timestamps of ratings. {Our tailored GP Model outperforms the na{\"i}ve GP, as indicated by the lower WAIC value observed in column~(3). Hence, including review heterogeneity variables beyond online ratings helps improve the model fit. Hence, we use the latter model for assessing the prediction performance on out-of-sample data at the deployment in the following section.

\subsection{Prediction Performance at Deployment Phase}

We tested the prediction performance of the estimated models in a separate deployment phase with out-of-sample data. For this, we collected additional ratings that were not used during estimation. At deployment time, all inferences are made \emph{without} having access to any review variables from the test set. Hence, the inferences are solely based on the past rating sequence (and past review heterogeneity variables). This ensures that we obtain a customer-independent score that could be displayed as part of an aggregated rating.

We then compared how well the different models inferred the expected rating. For this, we drew upon both the mean absolute error~(MAE) and the root mean square error~(RMSE). For each restaurant, the prediction was repeated for a hold-out set of ten subsequent ratings to remove the potential variability among individual ratings and ensure reliable performance results. Using other sizes for the hold-out set yielded consistent findings.

\Cref{tbl:benchmark_comparison} lists the results. The best out-of-sample performance is obtained by our tailored GP model. It registers an MAE of only 0.469, which is an improvement of \SI{10.4}{\percent} over the sample mean from industry. \Cref{tbl:benchmark_comparison} further reports a paired Wilcoxon signed rank test, confirming that the improvement over the sample mean is statistically significant.

It can be observed that the sample mean presents a strong baseline. The MAE is 0.523 stars. The other arithmetic baselines, despite their theoretical advantages, appear to be largely on par. A lower error was only observed for the machine learning baselines (\eg, the linear model, the random forest, and the neural networks). The performance of the LSTM is on par with that of the transformer. We emphasize that both the multivariate LSTM and the transformer are state-of-the-art models from deep learning and have access to exactly the same data as our tailored GP model, that is, rating scores and review heterogeneity variables. In comparison to both baselines from deep learning, the GP model appears to capture the dynamics behind ratings and their review heterogeneity variables in a more effective manner. As part of our comparisons, we use machine learning baselines that operate on the same data as the GP model. This is to test whether the performance gain from our tailored GP model is not only due to more covariates but also the result of better model specification. Again, we find consistent performance improvement, suggesting that our model specification is superior. 

\REV{Finally, the evaluation of the na{\"i}ve GP yields two interesting findings}. Recall that the na{\"i}ve GP is without a linear structure in the mean function and thus without any covariates. Here, no parameters are shared across restaurants, and, hence, this baseline model is fully unpooled, representing an entity-by-entity estimation.} First, the na{\"i}ve GP already achieves a large improvement over the sample mean (and other arithmetic baselines such as a weighted mean or a discounted mean). Evidently, the arithmetic baselines model the rating evolution via discrete time steps, while the na{\"i}ve GP accounts for actual timestamps and models the rating evolution in continuous time (e.g., a sample mean does not capture that time intervals between ratings become larger, which indicates a degradation in customer satisfaction). Hence, a large performance gain is due to how the time dimension between ratings is modeled. Second, we observe an improvement when extending a na{\"i}ve GP into our tailored, latent GP model, where we accommodate additional covariates in the GP mean function. Hence, the review heterogeneity variables are responsible for clear performance gains.

\begin{table}[H]
\OneAndAHalfSpacedXI
\centering
\sisetup{parse-numbers=false}
\caption{\REV{Entity-Level} Out-of-Sample Comparison.}
\label{tbl:benchmark_comparison}
{\footnotesize
\begin{tabular}{l cc}
\toprule
Model & \multicolumn{1}{c}{MAE} & \multicolumn{1}{c}{RMSE} \\
\midrule
\multicolumn{3}{l}{\textbf{Arithmetic baselines} (from marketing practice)} \\
Sample mean & 0.523 & 0.674 \\
Weighted mean & 0.521 & 0.645 \\	
Sliding window mean & 0.536 &  0.713 \\
Discounted mean & 0.529 & 0.674 \\
\midrule
\textbf{Machine learning baselines} & & \\
Linear model (feature engineering) & 0.504 & 0.652 \\
Linear model (covariates) & 0.536 & 0.688 \\
Random forest (feature engineering) & 0.507 & 0.659 \\ Random forest (covariates) & 0.513 & 0.673 \\
Univariate LSTM & 0.502 & 0.635 \\
Multivariate LSTM & 0.507  & 0.649 \\
Transformer & 0.506 & 0.639 \\
\midrule 
\textbf{GP-based models} & & \\
Na{\"ive} GP & $0.488^{***}$ &  $0.624^{***}$ \\
\quad \textit{Rel. improvement over best arithmetic baseline}& \textit{-6.3\,\%} & \textit{-3.5\,\%} \\[0.1cm]
\quad \textit{Rel. improvement over best machine learning baseline}& \textit{-2.8\,\%} & \textit{-1.7\,\%} \\[0.1cm]
Our tailored GP & $0.469^{***}$ & $0.602^{***}$ \\
\quad \textit{Rel. improvement over best arithmetic baseline}& \textit{-10.0\,\%} & \textit{-6.7\,\%} \\[0.1cm]
\quad \textit{Rel. improvement over best machine learning baseline}& \textit{-6.6\,\%} & \textit{-5.2\,\%} \\
\bottomrule
\multicolumn{3}{l}{Significance stars: $^{*}$ 0.05, $^{**}$ 0.01, $^{***}$ 0.001}
\end{tabular}
}
\end{table}

\subsection{\REV{Performance of Different Estimation Strategy}}

\REV{In our main analyses, we relied on MCMC for the estimation of our tailored GP. To assess the robustness and scalability of our framework, we additionally estimated the same GP model using the scalable variational inference (VI) algorithm introduced in \Cref{sec:estimation_strategies}.}

\REV{We first examine whether the use of our scalable VI compromises the out-of-sample performance of the LGPM (see \Cref{tbl:svi_comparison}). Accordingly, we use the above setting, fit our model using both full MCMC and our scalable VI, and then compute the prediction performance. Here, we find that fitting our model with scalable VI also achieves a large improvement over the sample mean, and is only slightly worse than with full MCMC. We further test statistically whether the difference in the RMSE is statistically significant, but must reject this hypothesis for common significance levels. This thus implies a similar overall prediction performance.}

\begin{table}[H]
\color{blue}
    \OneAndAHalfSpacedXI
	\centering
	\sisetup{parse-numbers=false}
	\caption{Comparison of full MCMC vs. scalable VI estimation.}
    \label{tbl:svi_comparison}
	{\footnotesize
		\begin{tabular}{l SSSS}
			\toprule
			& \multicolumn{4}{c}{Performance}  \\
			\cmidrule(lr){2-5}			
			Model & {MAE} & {RMSE} & {JSD} & {EMD}\\
			\midrule
			\textbf{Sample mean} & 0.523 & 0.674 & 0.100 & 0.603\\
			\midrule 
			\textbf{GP model} & & & & \\
			\quad Full MCMC & 0.469 &  0.602 & 0.092 & 0.555\\
			& (-10.325) & (-10.682) & (-8.000) & (-7.960) \\
			\quad Scalable VI & 0.489 & 0.616 & 0.093 & 0.567\\
			& (-6.500) & (-8.605) & (-7.000) & (-5.970) \\
			\bottomrule
			\multicolumn{5}{l}{Relative difference~(in \%) to sample mean in parenthesis.}
		\end{tabular}
	}
\end{table}

\begin{figure}[htb]
\centering
\caption{Computational Runtime (per Restaurant).}
\label{fig:svgp_runtime}
\includegraphics[width=.45\textwidth]{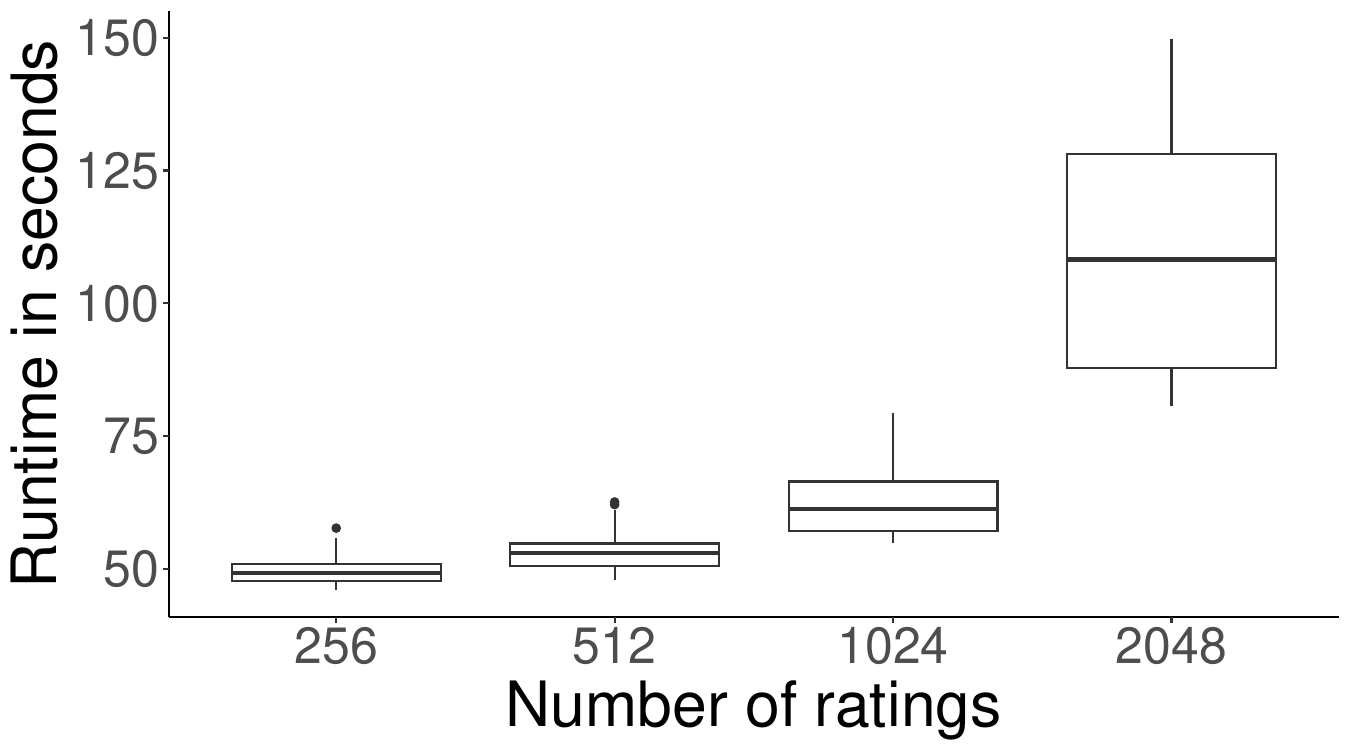}
\end{figure}

\REV{Next, we assess the computational efficiency of the scalable VI approach in \Cref{fig:svgp_runtime}. Each model was trained for 5,000 iterations on a standard desktop computer (equipped with an Intel i7-8550U (8th generation) CPU and 16 GB of RAM). Here, we matched restaurants to buckets that include up to $32$, $64$, etc. ratings (that is, by choosing the next larger bucket for a given number of reviews). We then computed the runtime at the restaurant level and reported the distribution over all per-restaurant runtimes as a boxplot. Even for longer rating sequences, the computational runtime increases only moderately. For comparison, full MCMC estimation for a restaurant with up to \REV{$256$} ratings already took around three minutes on the same machine and thus consumed more runtime by a factor of around three. For even larger datasets (such as those from the additional datasets with hotel ratings and app ratings), it was computationally intractable, and, for that reason, we only plot the scalable variational inference. In sum, this thus confirms the effectiveness of scalable variational inferences for our task and thus presents an estimation strategy that can be used for deploying the GP model at large platforms.}

\subsection{Additional Analyses}
\REV{
\subsubsection*{Sensitivity \REV{across rating volumes}.}

We analyzed how prediction performance varies with the number of available past ratings per restaurant (see Online Appendix~\ref{appendix:sens_length}). The results confirm that the tailored GP model consistently outperforms the sample mean across all rating volume segments. Interestingly, the relative improvement is most pronounced for restaurants with very short or very long rating histories.

\subsubsection*{\REV{Sensitivity across rating variation.}}

We further examined how predictive performance varies with the level of variation in prior ratings (see Online Appendix~\ref{appendix:sens_variation}). The analysis reveals that the improvement of our tailored GP model over the sample mean becomes more pronounced with higher prior rating variation. This pattern suggests that the GP’s ability to model temporal dynamics and uncertainty provides particular advantages when past ratings are volatile.

\subsubsection*{Results from additional datasets.}
\label{sec:results_additional_datasets}

To assess the generalizability of our findings, we evaluated our tailored GP model on two additional datasets, namely for hotel ratings and app ratings (see Online Appendix~\ref{appendix:add_datasets} for details). Across both datasets, our tailored GP consistently outperforms the sample mean and all machine learning benchmarks, confirming the robustness of our approach beyond the restaurant domain. While overall prediction errors are smaller for hotel ratings, the improvement over traditional aggregation methods remains statistically significant in both cases.

\subsubsection*{Practical Relevance of Results.}
\label{sec:pract_relevance}
\REV{To evaluate whether the above improvements in prediction performance translate into meaningful notable differences for consumers, we conducted two additional analyses (see Online Appendix~\ref{appendix:pract_implication}). First, we analyzed how accurate the rankings displayed in Yelp (with star ratings; not continuous values) would be as a result of our tailored GP model. This analysis reflects what users actually see and base their decisions on. We therefore compared the classification performance of displayed star ratings between the sample mean and our tailored GP model. Here, we found that our tailored GP model improves the accuracy of the predicted customer satisfaction by $11.979\%$ over the sample mean. Second, we simulated customer choice sets to examine how improved aggregation affects the ranking of restaurants and thereby the decision-making of customers. We again found a consistent improvement in the ranking accuracy of our tailored GP over the traditional mean-based aggregation. Importantly, the resulting restaurant rankings lead consumers to make substantially different choices across the two rating aggregation mechanisms, but where the GP framework helps users make decisions leading to greater expected satisfaction.}

\subsubsection*{Robustness Checks.}
\label{sec:robustness_checks}

We conducted a series of robustness checks to assess the stability of our results (see Online Appendix~\ref{appendix:robustness_checks} for full details). Across alternative model parameterizations, text-mining features, time periods, and kernel specifications, our findings remain consistent: the tailored GP model consistently outperforms all benchmark approaches. 
}

\section{Discussion}
\label{sec:discussion}

\subsection{Summary and Practical Implications}


Before making a purchase, customers want to learn about the quality of products or services to make informed decisions. In theory, customers can access various quality-related information on reputation systems (\eg, valence, helpfulness). However, people's cognitive abilities are limited, hindering them from processing large rating sequences \citep{Pope.2009}. This is especially true when they also have to take the heterogeneity of reviews or reviewers into account. Hence, many customers limit themselves only to the overall rating \citep{Tripadvisor.2017}. Taking these factors as motivation, we presented a data-driven framework for aggregating ratings that automatically interprets information available on reputation systems and condenses it into a single score in the form of an overall rating that is predictive of a customer's expected satisfaction. 


When comparing the aggregated score from the sample mean against the above GP model, two salient differences become evident. First, when using the sample mean, the heterogeneity among reviews or reviewers is ignored, while these data are fed directly into the GP model. Second, in the sample mean, old ratings are assigned the same weight as new ratings. Hence, if quality is subject to systematic changes, the average rating cannot catch up with the current quality level but remains subject to a \textquote{legacy bias} (see \Cref{fig:example_rating_aggregation} for an example). This issue occurs not only for services (\eg, restaurants) but also for physical products (\eg, books) whose perceived quality was found to be dynamic \citep{Li.2008}. Thus, for customers, the sample mean might be a misleading indicator of the current quality. For service providers, the sample mean can also be a burden because, when the sample mean is lower than the actual quality, customers might not make purchases. If the sample mean is higher than the current quality, improvements to the service (\eg, a better restaurant chef) are ineffective for the owner, as they will only be slowly picked up by the sample mean.


Our work facilitates the design of reputation systems in marketing applications. This is a crucial area of research, as only a few works have developed computational schemes for aggregating ratings. In practice, many reputation systems, such as those used by Amazon, Google Play, and Yelp, implement the sample mean (see \Cref{tbl:practice}). Our work highlights the limitations in current practice and provides an effective solution for rating aggregation. Notwithstanding, our framework is also relevant for traditional brick-and-mortar stores. Restaurants, for example, may want to show awards such as TripAdvisor's Certificate of Excellence on their front doors to attract customers, yet such awards are typically derived from reaching a certain overall rating.


Importantly, we explicitly refrained from personalizing the prediction to individuals. In many situations, personalization is unwanted, in particular, due to privacy reasons or when websites output a global rating score that is intentionally universal across the user base (\eg, comparison websites such as Yelp present universal scores). In other situations, personalization is infeasible, for example, when websites lack sufficient data (\eg, 71\,\% of the customers in our Yelp dataset made only one rating). In practice, most review platforms do not require users to ``sign in'', so they cannot offer individually customized ratings. Instead, they must rely on aggregated ratings. In line with this, our framework is designed for shopping websites or rating platforms where users are \emph{not} subject to tracking (\eg, IMDb, Rotten Tomatoes), where there is a cold-start problem (\eg, new customers without a purchase history), or where a universal score should be displayed to the entire user base (\eg, IMDb, Google Maps, Yelp, Amazon, TripAdvisor). This is an important difference from recommender systems, which rank products according to a customer's preferences \citep[\eg,][]{He.2020,Zhang.2012}. In contrast, we were interested in an aggregated score that is \emph{customer-independent} and that can be presented in reputation systems as an overall rating and thus to the entire customer base. Nevertheless, recommender systems could use the output of the GP model -- that is, our aggregated rating -- as an input to personalize recommendations. 

\subsection{Implications for Research}


\REV{In our work, we contribute to data-driven modeling for marketing research and online rating aggregation by empirically demonstrating the value of moving beyond simple averaging toward dynamic, data-driven aggregation. Our framework builds upon a tailored, latent GP model that naturally captures the temporal dynamics of ratings observed at irregular intervals and incorporates review heterogeneity through a structured mean function. While GPs have recently been introduced in marketing science \citep{Dew.2018,Dew.2020,Dew.2021,Dew.2024,Dew.2024b,Kim.2021}, they have not yet been tailored to the context of online ratings. Here, we present a novel empirical application, namely, rating aggregation.}

\subsection{Limitations and Potential for Future Research}


As with any research, our findings have limitations that open avenues for future research. First, while we considered many different model parameterizations, additional variables can be easily incorporated by updating the model parameterization. \REV{Second, further levels of unobserved, user-level heterogeneity could be incorporated, such as via user-level random effects within the GP mean function. However, in our empirical setting, no user provided more than one review per restaurant, and, hence, such an extension would not provide additional predictive value.} Second, both the mean function was intentionally formalized as a linear model to allow for interpretability. Nevertheless, the GP model outperforms alternative models that operate on the same data but with additional degrees of freedom due to nonlinearities (\ie, LSTMs from the area of deep learning). Third, the GP model could be extended so that different topics within the review text are considered. To achieve this, we expect the approaches described in \citet{Ansari.2018} and \citet{Toubia.2019} to be especially rewarding. Fourth, reputation systems may be subject to the strategic behavior of customers. To address this, we control for the average volume of ratings as well as the average rating score of a reviewer. Nevertheless, we acknowledge that accounting for such strategic behavior is a limitation inherent to essentially all reputation systems.

\section{Conclusion}
\label{sec:conclusion}


Many online marketplaces use reputation systems and, specifically, overall star ratings to inform customers. In this paper, we have shown that using the sample mean of past ratings for computing an overall star rating has shortcomings. \REV{To develop a better rating aggregation, we proposed the use of the GP framework, which is naturally dynamic and incorporates additional variables describing the heterogeneity in both reviews and reviewers. We empirically showed the predictive value of the GP using a comprehensive dataset of restaurant ratings and through a series of user studies. The results are valuable for marketers who want to condense data about customer satisfaction with products and services into an overall score.}





  \newcommand{\dq}{"}
	\renewcommand{\textquotedbl}{"}
\bibliographystyle{informs2014} 
{\OneAndAHalfSpacedXI
\bibliography{literature} 
}
\clearpage

\begin{APPENDICES}
\begin{center}
\Large\bfseries Online Appendix
\end{center}
\vspace{0.3cm}

\renewcommand\appendixname{Appendix}

\section{MCMC Estimation Details}
\label{appendix:estimation_details}

\subsection{Full MCMC Sampling from the Likelihood}
\label{sec:full_mcmc}


\REV{Full Markov chain Monte Carlo~(MCMC) estimates the model parameters \cite[see][]{Gelman.2014b}and, hence, enables} to directly sample from the posterior distribution of the model parameters. MCMC presents a highly flexible approach and is thus particularly suited for structural models with latent concepts. Another important benefit is that we later obtain credible regions for all model parameters, that is, all coefficients inside both the latent mean function and the latent covariance function. In the context of our work, this facilitates hypothesis testing. For instance, we can obtain significance levels at $p$\,\% by computing whether $(1-p)$\,\% of the HPDI (highest posterior density interval) does not contain 0. At this point, we emphasize that credible regions from MCMC estimation are robust against multicollinearity. 

For estimation, we first derived the log-likelihood $L$ for a given entity $i$, 
so that we can directly sample from it.
 Then we drew upon recent advances in Bayesian estimations \citep{Hoffman.2014}, namely, the Hamiltonian Monte Carlo~(HMC) sampler with the No-U-Turn~(NUTS) technique as implemented in Stan \citep{Carpenter.2017}. This approach differs from other estimation techniques such as the Metropolis-Hastings algorithm or maximum likelihood estimation. In contrast to them, our approach leverages an explicit derivation of likelihood in order to direct sample from the posterior distribution. This is known to be considerably more efficient and, together with Hamiltonian Monte Carlo, requires fewer chains/iterations by several orders of magnitude \citep{Hoffman.2014}. Owing to the efficiency of HMC over alternative samplers, two Markov chains, each with 2,500 iterations, were sufficient. The first 1,000 iterations of each chain served as a warm-up procedure and were subsequently discarded (note that, different from Metropolis-Hastings, no burn-in is required). The HMC sampler is computationally highly efficient: the analysis in this work was possible with standard office hardware, which, in settings with multiple cities, could be easily parallelized. We further applied reparameterization tricks from \citet{Rasmussen.2006}, which preserve the overall model expressiveness but improve convergence of the MCMC chains.


The estimation results were carefully checked against common guidelines \citep{Gelman.2014b}. Our sampling was informed by the so-called number of effective samples. Afterwards, convergence was ensured both by inspecting trace plots of chains and by computing the Gelman-Rubin convergence diagnostic. The latter ranges below the critical threshold of \num{1.02} for all parameters.

For completeness, the source code is reported in Online Appendix~\ref{appendix:source_code}.

\subsection{Prior Choice} 

We used weakly informative priors for all components inside our \REV{tailored} GP model as follows: 
\begin{enumerate}
\item \emph{Latent function.} Addressing high correlation in the latent function $f_i$, we used whitening for more efficient sampling. Specifically, we placed a standard normal prior on $\tilde{f}_i = L_i^{-1}f_i$, where $L_i$ is the Cholesky decomposition of the covariance matrix $K_i$. The entries of the covariance matrix are calculated via \Cref{eq:non_stationary_kernel}.
\item \emph{Emission component.} We chose a truncated standard Cauchy distribution~(centrality parameter equal to zero and scale parameter equal to one) for the standard deviation $\kappa_i$ of the Gaussian noise. Furthermore, we introduced a simplex vector $\eta_i\in\left(0,1\right)^{5}$ on which we placed a standard Dirichlet prior. The parameters $\eta_i$ were then transformed to the cut-points $c_{i(j)}$ for $j=1,\ldots,n_r-1$ by setting $c_{i(j)}=\kappa_i \, \mathrm{\Phi}^{-1}(\sum_{k=1}^j\eta_{ik})$. Thereby, one yields $\eta_{ik}=P(R_{ij}=k\mid f_{ij}=0)$. 
\item \emph{Mean function.} A QR-decomposition was applied to $x_i$, \ie, $x_i = Q_i\,R_i$. We then placed a standard normal prior on the parameter $\tilde{\theta} = R_i\theta$. 
\item  \emph{Kernel.} We placed a truncated standard normal prior on the kernel standard deviation $\sigma_i$, so that we allow for a potentially small correlation between ratings. For the length scale parameter $\rho_i$, we chose a more informative prior in order to ensure efficient sampling: we drew upon the temporal distance between ratings, $\Delta_{jk}(i) =\abs{t_{ij}-t_{ik}}$, to restrict the length scale to an interval between the minimal and maximal temporal distance. Formally, we computed $l_i=\min_{j,k=1,\ldots,n_i}\Delta_{jk}(i)$  and $u_i = \max_{j,k=1,\ldots,n_i}\Delta_{jk}(i)$. Based on it, we restricted the prior in a way that only around \SI{1}{\percent} of prior probability mass is outside the interval $\left[l_i,u_i\right]$. {This was achieved by choosing an inverse gamma function as the prior and using an algebraic solver to find the shape and scale parameters, so that the above condition was met.} 
\end{enumerate}

\REV{
\subsection{Posterior Summaries}\label{apx:post_summs}

We summarize the posterior distributions obtained from the MCMC estimation of our tailored GP model. Specifically, we report the posterior medians and 95\% credible intervals for the parameters inside the kernel, namely, $\rho_i$ and $\sigma_i$ (see \Cref{fig:post_summs}). These posterior summaries allow us to verify that the posterior intervals are well-behaved and centered within reasonable ranges of the corresponding priors.

\begin{figure}[htb]
\centering
\caption{Posterior Summaries for Entity-Specific Parameters inside the Kernel.}
\label{fig:post_summs}

\begin{minipage}{0.48\textwidth}
    \centering
    \includegraphics[width=\linewidth]{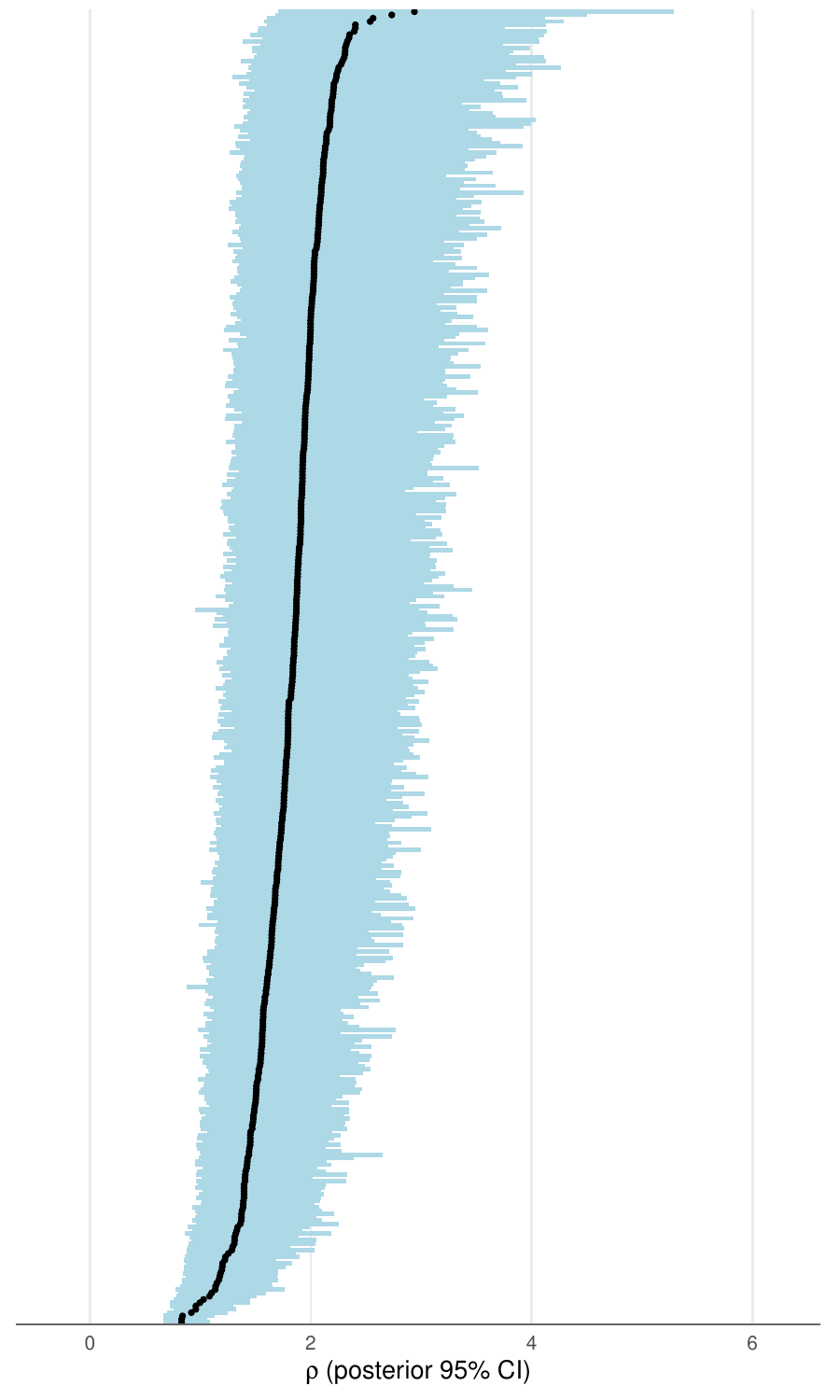}
\end{minipage}
\hfill
\begin{minipage}{0.48\textwidth}
    \centering
    \includegraphics[width=\linewidth]{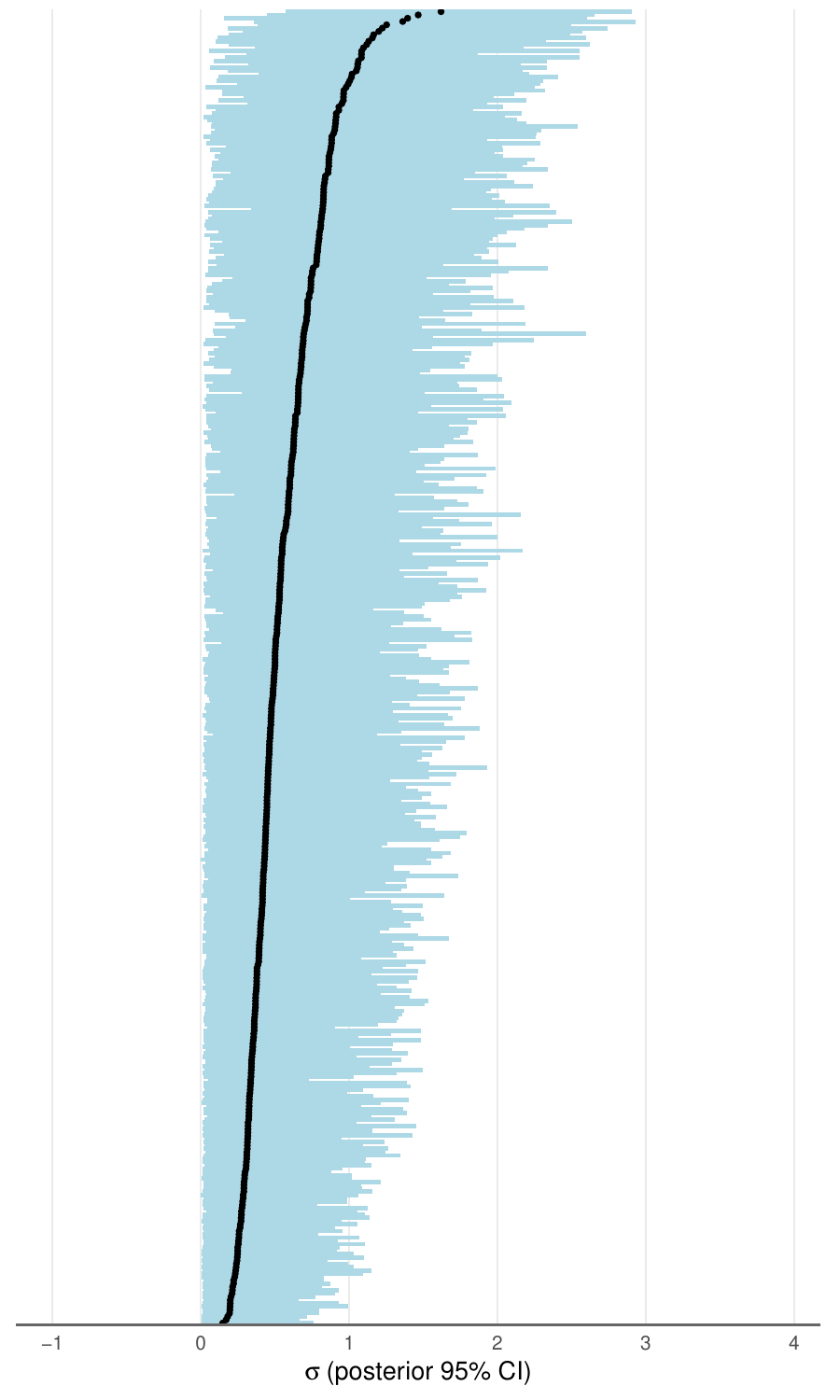}
\end{minipage}

\end{figure}

\subsection{Parameter Recovery}\label{apx:parameter_recovery}

We also ran additional robustness checks concerning our model specification in order to ensure that we can retrieve the original parameters from simulated data. Specifically, we generated a simulated dataset using the same GP structure and ordinal outcome model as in the main analysis, with known ground-truth parameters ($\theta$, $\sigma_i$, $\rho_i$). The simulated dataset contained 8 entities, each with 80 reviews and two covariates governing the mean function through true coefficients $\theta=(0.0,0.1)$. For every entity $i$, we drew an individual GP length-scale and variance parameter from uniform distributions, 
$\rho_i\sim U(0.7,1.2)$ and $\sigma_i\sim U(0.9,1.3)$, respectively. These parameters defined entity-specific latent functions with exponential covariance kernels, from which ordinal outcomes were simulated using a fixed set of balanced cutpoints. We then refitted our tailored GP model on the simulated dataset.

We evaluated the recovery of the simulated parameters based on three standard metrics. For each parameter ($\theta$, $\sigma_i$, $\rho_i$), we computed the bias, the RMSE, and the empirical coverage rate to check whether the true value fell within its 95\% posterior interval. Table \ref{tbl:recovery} summarizes the results of the parameter-recovery exercise. Overall, the model recovers the simulated parameters accurately. The mean function coefficients ($\theta$) show negligible bias and perfect empirical coverage, confirming that the main regression structure is well identified. The entity-specific GP parameters ($\rho_i$ and $\sigma_i$) are also well recovered: both exhibit small bias and low RMSE, with coverage rates close to the nominal 95\%. The slightly lower coverage for $\rho_i$ (83\%) reflects mild underestimation of posterior uncertainty for a few entities but remains within acceptable bounds for hierarchical GP settings with limited per-entity data.

\begin{table}[ht]
\OneAndAHalfSpacedXI
\centering
\sisetup{parse-numbers=false}
\caption{Parameter Recovery} 
\label{tbl:recovery}
\begin{tabular}{lrrr}
  \hline
Parameter & Bias & RMSE & Coverage (95\%) \\ 
  \hline
$\theta$ & -0.03 & 0.09 & 1.00 \\ 
  $\rho_i$ & -0.10 & 0.15 & 0.83 \\ 
  $\sigma_i$ & 0.06 & 0.36 & 1.00 \\ 
   \hline
\end{tabular}
\end{table}

}

\section{Proof of Theorem 1} 
\label{appendix:proof}

The derivation follows directly from \citet{Hensman.2015}. We denote the realization of the latent function $f_i$ at $M$ as $u$. Then we derive a first lower bound for the log likelihood of the observed rating sequence using Jensen's inequality via
\begin{equation}
\label{eq:elbo_1}
\log p(Y_i)\geq\mathbb{E}_{q(u_i)}\lbrack\log p(Y_i \,\mid\, u_i)\rbrack - \mathrm{KL}\lbrack q(u_i) \,\mid\mid\, p(u_i) \rbrack.
\end{equation}
Furthermore, we derive an additional inequality for $\log p(Y_i \,\mid\,u_i)$, again using Jensen's inequality. This yields
\begin{align}
\log p(Y_{i} \,\mid\,u_i) &=\log\int p(Y_{i},f_i \,\mid\,u_i) \dd f_i \\
&=\log\int\frac{p(Y_{i} \,\mid\, f_i) \, p(f_i,u_i)}{p(u_i)} \dd f_i \\
&\geq\mathbb{E}_{p(f_i \,\mid\, u_i)}\lbrack\log p(Y_i \,\mid\, f_i)\rbrack.
\end{align} 
Inserting the previous result into \Cref{eq:elbo_1} yields the final ELBO:
\begin{equation}
\label{eq:elbo_final}
\log p(Y_i)\geq\mathbb{E}_{q(f_i)} \lbrack\log p(Y_i \,\mid\, f_i)\rbrack - \mathrm{KL}\lbrack q(u_i) \,\mid\mid\, p(u_i) \rbrack,
\end{equation}
where $q(f_i) = \int p(f_i \,\mid\, u_i) \, q(u_i) \dd u_i$. This thus proves Theorem~1. 

\section{Implementation of Baselines}
\label{appendix:baselines_implementation}

The baselines are implemented as follows:
\begin{quote}\begin{itemize}
\item \textbf{Sample mean:} It simply averages over all ratings via $\frac{1}{n_i}\sum_{j=1}^{n_i}R_{ij}$. 
\item \textbf{Weighted mean:} Formally, it models the expected value via a \textquote{true Bayesian estimate} \citep{IMDb.2025}, that is, averaging over all discrete ratings with respect to their frequency (\ie, Bayesian averaging). The model is estimated by assuming a categorical distribution for the ratings $R_{i,1},\ldots,R_{i,n_i}$.
\item \textbf{Sliding window mean:} Here, the mean is computed based on the last $l_i$ ratings of each sequence, \ie, $\frac{1}{l_i}\sum_{j=n_i-l_i+1}^{n_i}R_{ij}$ \citep{Ivanova.2017}. The restaurant-specific parameter $l_i$ was determined via time-series cross-validation with the size of the hold-out data sets fixed to \num{5} ratings. 
\item \textbf{Discounted mean:} Older ratings are discounted by placing a weight $\omega_{ij}$ on each rating \citep{Leberknight.2012}. Formally, the expected value is given by $\sum_{j=1}^{n_i}\omega_{ij}R_{ij}$ with weights $\omega_{ij}=\exp\left(-\lambda_i\left(n_i-j+1\right)\right)$. Here, $\lambda_i$ was tuned via the above time-series cross-validation.
\item \textbf{Long short-term memory (LSTM):} We use the LSTM as a state-of-the-art recurrent neural network (RNN). The LSTM decomposes the likelihood of an observed rating sequence as 
\begin{equation}
P\left(R_{i,1},\ldots,R_{i,n_i}\right)=\prod_{k=1}^{n_i}P\left(R_{i,k}\mid R_{i,1},\ldots,R_{i,k-1}\right).
\end{equation}
The likelihood of each rating $R_{i,k}$ is conditioned on the prior ratings through a latent vector $\bm h_{i,k-1}$ which depends both on $\bm h_{i,k-2}$ and the rating $R_{i,k-1}$ through a deterministic function $H$, the so-called LSTM unit \citep{Hochreiter.1997}. This yields
\begin{align}
\bm h_{i,k} &= H\left(\bm h_{i,k-1},R_{i,k}\right) \qquad\qquad\quad \text{ and}\\
P\left(R_{i,k}\mid R_{i,1},\ldots,R_{i,k-1}\right) &= \textup{softmax}\left(\bm W\bm h_{i,k-1}+\bm b\right) 
\label{eq:lstm_prob}.
\end{align}
Here, the matrix $\bm W$ links the latent vector to the rating. The expected rating $\mathbb E\left[R \mid \cdot \right]$ is calculated using the probability distribution in \Cref{eq:lstm_prob}, that is, $\textup{softmax}\left(\bm W\bm h_{i,n_i}+\bm b\right)$. The architecture was analogously extended to handle multivariate input.

In our experiments, the network architecture consisted of a single LSTM layer followed by a fully-connected layer. We also experimented with multi-layer architectures but which led to overfitting. The output dimension of the LSTM layer was fixed to \num{50} neurons; the dimension of the fully-connected layer was set to \num{5} neurons corresponding to the five different star ratings. {Activation functions were set to hyperbolic tangent, expect for a softmax in the fully-connected layer. A random dropout of $0.5$ given the considerable size of the parameter space. To ensure a fair comparison and handle the between-entity heterogeneity, we fitted an individual LSTM to each rating sequence. Training was done for \num{10} epochs using the Adam optimizer with the default learning rate of \num{0.001}.} 

{We also considered a bidirectional LSTM but, due to the larger parameter space, its performance was found to be inferior and, for reasons of brevity, the corresponding results were omitted. When implementing different forms of attention, similar observations were made. }
\item \textbf{Transformers:} Pre-trained transformers are widely used for many applications that involve, for instance, natural language processing \citep[\eg,][]{Liu.2020,Wolf.2020}. For our application, such a pre-trained transformer is not available. Instead, we train our own transformer network but follow the architecture from \citet{Vaswani.2017} with the only exception that we replace the classification by a regression task. In brevity, it consists of a multi-head attention with dropout (0.1). In addition, the feed-forward part is performed by fusing a 1-dimensional convolutional layer with kernel size of 1 and ReLU activation, followed by dropout (0.1), and another 1-dimensional convolutional layer. The final prediction is done via global average pooling through a dense layer with ReLU activation and droput (0.2). The transformer has a comparatively large number of parameters, which quickly leads to overfitting when additional hyperparameter tuning is applied. Instead, we only slightly vary the default hyperparameters from Keras and begin with the number of heads set to 4, the head size to 256, and use 4 transformer blocks. Details are in \url{https://keras.io/examples/timeseries/timeseries_transformer_classification/}.
\end{itemize}
\end{quote}
\noindent
Hyperparameter for tuning the baseline models are listed in \Cref{tbl:ml_model_tuning}.

\begin{table}[H]
	\centering
	\SingleSpacedXI
	\caption{Hyperparameters for Tuning the Baseline Models.\label{tbl:ml_model_tuning}}
	{\footnotesize
		\begin{tabular}{l ll}
			\toprule
			{Baselines} & {Hyperparameters} & Tuning range \\
			\midrule
			Sample mean & None & -- \\
			Bayesian averaging & Concentration parameter $\alpha_i$ & $0.1,0.25,0.5,1,2,4,8$ \\
			Sliding window mean & Window size $l_i$ & $1,\ldots,n_i-5$ \\
			Discounted mean & Discount rate $\lambda_i$&$0,0.5,1,2,4,8,16$  \\
			LSTM & Size of state vector & $5,10,25,50,100$ \\
			& Activation function & Rectified Linear Unit,Soft Plus, Hyperbolic Tangent \\
			& Dropout rate &$0.0,0.1,0.25,0.5,0.75$ \\
			& Learning rate &$0.0001,0.001,0.01,0.1$ \\
			Transformer & Number of attention Heads & $3,4,5,6$\\
			& Size of attention Heads & $64,128, 256, 512$\\
			& Convolution filter size & $2,4, 8$\\
			& Number transformer blocks & $2, 4, 6$\\
			& Size of MLP & $16, 32, 64, 128, 256$\\
			\bottomrule
		\end{tabular}
	}
	\footnotesize
	\begin{tabular}{p{16cm}} \emph{Notes:} Hyperparameters for each of the baselines models and the corresponding values considered during tuning via five-fold time-series cross-validation. \end{tabular}
\end{table}


\section{Additional Analyses}

\subsection{Sensitivity across Rating Volumes}\label{appendix:sens_length}

\Cref{tbl:range_performance} reports the results of a sensitivity analysis in which we compare the prediction performance across restaurants with different rating volumes. This validates that the improvement of our \REV{tailored} GP model over the sample mean is robust. The larger relative improvement for short and long rating sequences can be explained as follows. Long rating sequences are often linked to established restaurants that have been around for a long time and that have thus a good reputation and where rating sequences thus have less variance. For instance, the average standard deviation in rating scores for the $\geq 500$ bucket amounts to \num{1.09}, while it amounts to \num{1.28} for the 50--99 bucket, explaining why predictions for long sequences are less challenging. Similarly, new restaurants with short rating sequences are also characterized by a lower variance (\ie, \num{1.33} for the 25--49 bucket), explaining why predictions are less challenging.

\begin{table}[H]
	\OneAndAHalfSpacedXI
	\centering
	\sisetup{round-mode=places,round-precision=3}
	\caption{Sensitivity Analysis by Rating Volume. The restaurants are grouped into buckets by the different number of available ratings.} \label{tbl:range_performance}
	{\footnotesize
		\begin{tabular}{rr SS SS r}
			\toprule
			&  & \multicolumn{2}{c}{Sample mean} & \multicolumn{2}{c}{Our \REV{tailored} GP}  \\
			\cmidrule(lr){3-4}			\cmidrule(lr){5-6}
			{Bucket} & {\#} & {MAE} & {RMSE} & {MAE} & {RMSE} & Rel. improvement in MAE \\			
			\midrule		
  $<$ 25 & 125 & 0.673877790303204 & 0.82591472157065 & 0.568965495407301 & 0.707055859030478 & -15.57 \\ 
  25 -- 100 & 195 & 0.458420053242842 & 0.606120007142063 & 0.426159195753906 & 0.554887029246644 & -7.04 \\ 
  100 -- 200 & 91 & 0.438748241466105 & 0.552497558115383 & 0.406502450977595 & 0.51737317897568 & -7.35 \\ 
  200 -- 500 & 163 & 0.330938116492026 & 0.423419590849468 & 0.316317879447853 & 0.402164942726023 & -4.42 \\ 
  $\geq$500 & 51 & 0.263810763749409 & 0.333189633899522 & 0.234600875098039 & 0.297697163996421 & -11.07 \\ 
			\bottomrule
			\multicolumn{6}{l}{\# $=$ number of restaurants}
		\end{tabular}
	}
\end{table}

\REV{
\subsection{Sensitivity across Rating Variation}\label{appendix:sens_variation}

\REV{\Cref{tbl:variation_performance} reports the results of a sensitivity analysis in which we compare the prediction performance across restaurants with different rating variations. This provides more insights into where our tailored GP model achieves the highest performance improvement over the sample mean, validating that the improvement of our tailored GP model over the sample mean is robust.}

\begin{table}[H]
	\OneAndAHalfSpacedXI
	\centering
	\sisetup{round-mode=places,round-precision=3}
	\caption{Sensitivity Analysis by Rating Variation. The restaurants are grouped into buckets by the level of prior rating variation.}\label{tbl:variation_performance}
    \REV{
	{\footnotesize
		\begin{tabular}{rr SS SS r}
			\toprule
			&  & \multicolumn{2}{c}{Sample mean} & \multicolumn{2}{c}{Our tailored GP}  \\
			\cmidrule(lr){3-4}			\cmidrule(lr){5-6}
			{Bucket (in std dev)} & {\#} & {MAE} & {RMSE} & {MAE} & {RMSE} & Rel. improvement in MAE \\			
			\midrule	
 $<$ 1.0 & 99 & 0.247532117056686 & 0.308917133876993 & 0.257857650200105 & 0.326397230728146 & 4.17\,\% \\ 
  1.0--1.2 & 123 & 0.337870299403778 & 0.441065020752009 & 0.322577574385035 & 0.419906855168127 & $-$4.53\,\% \\  
  1.2--1.3 & 101 & 0.443254660249951 & 0.575289487899757 & 0.400085112169139 & 0.54186473487552 & $-$9.74\,\% \\ 
  1.3--1.5 & 196 & 0.497993381192103 & 0.624968395940381 & 0.439136108201355 & 0.551727152448593 & $-$11.82\,\% \\ 
  $\geq$1.5 & 106 & 0.684064493102985 & 0.84792847453837 & 0.594846127935541 & 0.728269158719671 & $-$13.04\,\% \\ 
			\bottomrule
			\multicolumn{6}{l}{\# $=$ number of restaurants}
		\end{tabular}
	}
    }
\end{table}

}
\subsection{Results from Additional Datasets}\label{appendix:add_datasets}

We evaluate the tailored GP model using two additional datasets with ratings, namely, (1)~hotel ratings and (2)~app ratings.\footnote{\SingleSpacedXI\footnotesize We used scalable variational inferences for efficient estimation (see \Cref{sec:estimation_strategies}).} For the former, we used \num{53217} ratings from \citet{Mattei.2017}, and for the latter, \num{153} apps from the Google Play store with overall \num{114344} ratings.\footnote{\SingleSpacedXI\footnotesize This dataset can be obtained at kaggle.com: \url{https://www.kaggle.com/datasets/mehdislim01/google-play-store-apps-reviews-110k-comment?select=Apps.csv}} We used the same data processing (\eg, the same train/test split, the same hyperparameter tuning) and the same baselines as above.

The results are shown in \Cref{tbl:benchmarks_additional_datasets}. For both datasets, we find that our \REV{tailored GP} model has superior performance. It again outperforms both the sample mean used in current practice as well as all machine learning baselines. When comparing both datasets, we note a tendency for overall errors to be smaller for hotel ratings compared with app ratings. We observe this tendency for all aggregation mechanisms. One reason is that app updates can change the underlying quality of the app, and the models need a short time frame to adapt. The table further reports the results from the paired Wilcoxon signed rank test, confirming that the improvement over the sample mean is statistically significant.

\begin{table}[H]
\OneAndAHalfSpacedXI
\centering
\sisetup{parse-numbers=false}
\caption{Out-of-Sample Comparison.}
\label{tbl:benchmarks_additional_datasets}
{\footnotesize
\begin{tabular}{l SS SS}
\toprule		
Model & \multicolumn{2}{c}{Hotel ratings} & \multicolumn{2}{c}{App ratings} \\
\cmidrule(lr){2-3} \cmidrule(lr){4-5}	\\
& \multicolumn{1}{c}{MAE} & \multicolumn{1}{c}{RMSE} & \multicolumn{1}{c}{MAE} & \multicolumn{1}{c}{RMSE} \\
\midrule
\multicolumn{5}{l}{\textbf{Arithmetic baselines} (from marketing practice)} \\
Sample mean & 0.378 & 0.479 & 0.523 & 0.728 \\
Weighted mean & 0.400  & 0.519 & 0.528 & 0.724   \\	
Sliding window mean & 0.395  & 0.512 & 0.498 & 0.672 \\
Discounted mean &0.367  & 0.469 & 0.581 & 0.756\\
\midrule
\textbf{Machine learning baselines} & & & & \\
Linear model (feature engineering) & 0.372 & 0.471 & 0.537 & 0.726 \\
Linear model (covariates) & 0.371 & 0.470 & 0.534 & 0.719\\ 
Random forest (feature engineering) & 0.370 & 0.469 & 0.487 & 0.652 \\
Random forest (covariates) & 0.390 & 0.492 & 0.561 & 0.744 \\
Univariate LSTM & 0.371 &0.466 & 0.496 & 0.665 \\
Multivariate LSTM & 0.368  & 0.465& 0.489 & 0.643  \\
Transformer & 0.365 & 0.472 & 0.511 & 0.639\\
\midrule 
\textbf{GP-based models} & & & & \\
Na{\"ive} LGPM & 0.361 & 0.465 & 0.488 & 0.641 \\
\REV{Our tailored} GP & 0.353^{***} & 0.456^{***} & 0.476^{***} & 0.626^{***} \\
\bottomrule
\multicolumn{5}{l}{Significance stars: $^{*}$ 0.05, $^{**}$ 0.01, $^{***}$ 0.001}
\end{tabular}
}
\end{table}

\REV{
\subsection{Practical Relevance of Results}\label{appendix:pract_implication}

To illustrate the practical impact of the tailored GP compared to traditional mean-based aggregation, we evaluated the predictions of our tailored GP model as a classification task (see \Cref{tbl:classification_comparison}). Specifically, we rounded the predictions from our model to the nearest star as displayed on platforms such as Yelp. Again, our tailored GP outperforms the mean-based aggregation across all performance metrics. For instance, the balanced accuracy is improved by $11.979\%$.

\begin{table}[H]
    \OneAndAHalfSpacedXI
	\centering
	\sisetup{parse-numbers=false}
	\caption{Out-of-sample Classification Performance.}
    \label{tbl:classification_comparison}
	{\footnotesize
		\begin{tabular}{l SSSS}
			\toprule
			& \multicolumn{4}{c}{Performance}  \\
			\cmidrule(lr){2-5}			
			Model & {Balanced accuracy} & {Balanced F1} & {Balanced precision} & {Balanced recall}\\
			\midrule
			\textbf{Sample mean} & 0.480 & 0.467 & 0.475 & 0.480\\
			\textbf{Our tailored GP} & 0.538 & 0.519 & 0.539 & 0.538 \\
            \midrule
			\textbf{Rel. improvement~(in \%)} & -11.979 & -11.231 & -13.478 & -11.979 \\
			\bottomrule
		\end{tabular}
	}
\end{table}

To further illustrate the practical implications of our model, we simulated customer decision-making based on predicted restaurant ratings. Specifically, we constructed random choice sets of restaurants and compared how often the restaurant with the highest true satisfaction was correctly identified under different aggregation methods. We performed this simulation for choice set sizes of $K =$ 5, 7, and 10, averaging results over 2,000 randomly generated choice sets to ensure stability. For each setting, we computed Top-1, Top-2, and Top-3 accuracies, i.e., the proportion of cases in which the restaurant with one of the highest true ratings would have been correctly recommended. We again find a consistent improvement in the ranking accuracy of our tailored GP over the traditional mean-based aggregation (see \Cref{tbl:choice_set}).

\begin{table}[H]
    \OneAndAHalfSpacedXI
	\centering
	\sisetup{parse-numbers=false}
	\caption{Simulated Choice Set Ranking Performance.}
    \label{tbl:choice_set}
	{\footnotesize
		\begin{tabular}{l SSS}
			\toprule
			& \multicolumn{3}{c}{Performance}  \\
			\cmidrule(lr){2-4}			
			\quad Model & {Top-1 accuracy} & {Top-2 accuracy} & {Top-3 accuracy}\\
			\midrule
            \textbf{$K=5$} &  & & \\
			\quad\textbf{Sample mean} & 0.476 & 0.645 & 0.772\\
			\quad\textbf{Our tailored GP} & 0.518 & 0.676 & 0.790 \\
            \midrule
            \textbf{$K=7$} &  & & \\
            \quad\textbf{Sample mean} & 0.422 & 0.570 & 0.666 \\
			\quad\textbf{Our tailored GP} & 0.441 & 0.594 & 0.686 \\
            \midrule
            \textbf{$K=10$} &  & & \\
            \quad\textbf{Sample mean} & 0.337 & 0.478 & 0.571\\
			\quad\textbf{Our tailored GP} & 0.358 & 0.513 & 0.613 \\
			\bottomrule
		\end{tabular}
	}
\end{table}
}
\subsection{Robustness Checks}\label{appendix:robustness_checks}

\REV{
We varied the time period under study and focused only on ratings from 2016 and 2017. The use of a shorter time period could theoretically result in the removal of structural changes in the data, which is an advantage for the sample mean; however, such behavior is already formalized in the sliding window mean, which focuses only on recent ratings. For our tailored GP model, changing the time period also led to a similar performance. In summary, we obtained conclusive findings. 

The number of helpfulness votes can be computed either at the review or at the reviewer level. In our evaluation, we followed \citet{Weiss.2008} and used helpfulness at the reviewer level. To justify this choice, we also performed a robustness check in which we replaced the number of helpfulness votes at the reviewer level with the number of helpfulness votes at the review level. We then compared both to determine which model specification appears more beneficial. Across all performance metrics (WAIC, RMSE, and MAE), the model variant with reviewer helpfulness is better than the model with review helpfulness.
}

We further experimented with alternative model parameterizations. First, we replaced the time on Yelp as a measure of experience by the overall number of reviews by an individual. However, this did not improve the model fit. A potential reason is that users undergo periods without reviewing but reading reviews, so that time is better reflecting experience. Furthermore, we accounted for a potential social influence, in which a user is influenced by prior ratings (\ie, with autoregression) \citep{Xie.2023}. However, this was not beneficial, as the interpolation directly accommodates past ratings.

\REV{
We also experimented with other review textual features, specifically term frequencies \citep[\eg,][]{Li.2011b,Ling.2014,Wang.2016,Feuerriegel.2025}. A potential benefit of including term frequencies (or \mbox{tf-idf}) is that we control for the raw content in the written review; however, we receive a much larger set of predictors ($p \gg 100$). We found that this led to overfitting, even when applying dimensionality reduction. As a solution, we followed the feature engineering literature and controlled for the positivity/negativity of written reviews by extracting the review sentiment \citep[\eg,][]{Naumzik.2021,Tirunillai.2012}. For future work, it would be interesting to extend our model with other text mining frameworks for topic analysis \citep{Zhong.2020}.

In addition to the exponential kernel, we considered a wide range of covariance functions, including a squared exponential (SE) kernel and a rational quadratic kernel. The intuition behind this is that ratings may either shift quickly over time due to abrupt quality changes, for example, when a chef changes (captured via the exponential kernel) or remain more stable over time (SE kernel). The exponential kernel would allow for frequent changes in quality, while, under the SE or the rational quadratic kernel, ratings remain correlated for a longer time (old reviews have a stronger influence). Both -- SE kernel and rational quadratic kernel -- impose strong smoothness assumptions on the function $f_i$ \citep[Ch.\,4]{Rasmussen.2006}; however, in our experiments, these smoothness assumptions appeared unrealistic, and the corresponding models generally resulted in an inferior fit. The exponential kernel we employ is stationary, meaning that the covariance function depends on the temporal distance $t_{ij}-t_{ik}$ between ratings. This implies that the relationship between ratings remains the same across the entire time domain. Here, stationarity simplifies modeling and improves interpretability. We also experimented with more flexible kernels (e.g., nonstationary exponential kernels). Our results indicate that, while a nonstationary kernel structure could improve local flexibility, the stationary exponential kernel provided a favorable trade-off that achieves strong predictive performance and maintains interpretability.
}


Our evaluation was repeated with alternative performance measures that compare the posterior rating distribution and the empirical distribution. Here, we used the Jensen-Shannon divergence and the Earth mover's distance. For our GP model, the posterior rating distribution is calculated via \Cref{eq:expectation}. For the sample mean, we simply use the empirical distribution of the training sequence of each restaurant. For the sliding window average, we only consider the empirical distribution of the ratings within the respective windows. For the LSTM, the posterior distribution is retrieved from the final softmax layer. The results are consistent: our GP model still provides the best-performing model by a substantial margin. In addition to an LSTM, we implemented a bidirectional LSTM (so-called BiLSTM), yet finding that it does not improve over the standard LSTM.

\section{Comparison to Computer Science Baselines}
\label{appendix:baselines_cs}

We now compare our tailored GP model against additional baselines from computer science. Specifically, we study baselines from other tasks, namely, (1)~recommender systems, (2)~information aggregation (``crowdsourcing''), and (3)~truth discovery. These tasks show parallels to rating aggregation but also have important differences (\eg, baselines for crowdsourcing tasks assume a static setting but without temporal dynamics as we do). As we shall we below, compared to state-of-the-art baselines from tasks (1)--(3), our tailored GP model is superior.

\subsection{Baselines from Recommender Systems}

Recommender systems perform information filtering with the objective of personalizing the result to the preference of users \citep[\eg,][]{Bhargava.2015,Elkahky.2015,Zhang.2012}. Speficially, we considered several state-of-the-art algorithms such as, \eg, collaborative filtering \citep{Schafer.2007}, matrix factorization \citep{Mnih.2007}, and hidden factors models \citep{McAuley.2013}. However, there are crucial differences that render algorithms from recommender systems inapplicable for our task. In recommender systems, ratings represent usually the input, while the output is a ranking of items in form of a ordered list (\ie, $k \in \{ 1, \ldots, N \}$), but not the expected value of the star-rating (\ie, $r\in\mathbb{R} $). However, both represent different tasks: ranking vs. regression. Because of that, both operate on different scales -- ordered lists and numerical scores. Yet these do not allow for comparison and, therefore, do not allow for performance benchmarking. 

On top of that, recommender systems from practice are intended to perform such rankings by considering a user's past preferences \citep[\eg,][]{Kahng.2011,Karatzoglou.2010}. Yet such an approach contradicts the intention of rating aggregation: aggregated ratings should neither provide a ranking nor a personalized result but an overall score that is \emph{independent} of a specific user. 

\subsection{Baselines from Crowdsourcing}


Crowdsourcing aims at a specific form of information aggregation: worker-specific performance scores are estimated from individual performance scores of works across a set of heterogeneous tasks with varying levels of difficulty \citep{Dawid.1979,Paun.2018,Zhang.2014b}. As such, it determines the ``true'' label for an entity based on multiple heterogeneous labels from crowd workers. Noteworthy is the seminal work by \citet{Dawid.1979}. Later efforts have been concerned with, for instance, provably-optimal \citep{Zhang.2014b} and hierarchical \citep{Paun.2018} implementations.


Crowdsourcing thus shares a similarity with rating aggregation, as it infers an overall performance score for an entity from individual values. However, algorithms for crowdsourcing differ from rating aggregation in several ways: (1)~Crowdsourcing assumes the worker-specific performance to be constant over time, while, in rating aggregation, we known that quality is dynamic \citep{Li.2008}. (2)~Crowdsourcing ignores the ordering of performance scores, whereas, in our task, ratings are ordered by time. (3)~Crowdsourcing considers only performance scores but no additional information. In contrast, our task of rating aggregation accommodates additional variables from reputation systems (\eg, helpfulness, elite status). However, such variables are absent in crowdsourcing.


In our experiments, we use the following state-of-the-art algorithms. Specifically, we implemented the original Dawid-Skene algorithm \citep{Dawid.1979}. We also use several extensions of it, namely, for provably-optimal crowdsourcing \citep{Zhang.2014b}, the logistic random effects model \citep{Paun.2018}, and the hierarchical Dawid-Skene algorithm \citep{Paun.2018}. Crowdsourcing algorithms rely on worker-individual parameters and thus require multiple observations for each worker. However, many raters in our dataset provide only a single review, and, therefore,
inferences of individual-level performance scores via the Dawid-Skene algorithm become numerically challenging. This shortcoming is theoretically addressed by the hierarchical model which pools rater-individual parameters.


\Cref{tbl:comparison_crowdsourcing} reports the results. Compared to our tailored GP model, we find that the crowdsourcing baselines are inferior by a considerable margin. For example, the MAE from the Dawid-Skene algorithm \citep{Dawid.1979} is more than \SI{100}{\percent} larger than that of the LGPM. The MAE for the provably-optimal variant \citep{Zhang.2014b} is better but still larger by a two-digit percentage. In case of the hierarchical Bayesian models \citep{Paun.2018}, we find that the results are also unfavorable.

\begin{table}
	\SingleSpacedXI
	\centering
	\sisetup{parse-numbers=false}
  \caption{Comparison Against Crowdsourcing Baselines.}
	\label{tbl:comparison_crowdsourcing}
	{\footnotesize
		\begin{tabular}{l SS}
			\toprule	
			Model & \multicolumn{1}{c}{MAE} & \multicolumn{1}{c}{RMSE} \\ 
			\midrule
			Dawid-Skene~(DS) algorithm \citep{Dawid.1979} & 1.095 & 1.345\\ 
			Provably-optimal crowdsourcing \citep{Zhang.2014b} & 0.639 & 0.717\\ 
			Logistic random effects model \citep{Paun.2018} & 0.642 & 0.778\\ 
			Hierarchical DS algorithm \citep{Paun.2018} & 1.651 &  1.865\\ 
			\midrule
			Our tailored GP & 0.470 & 0.612 \\			
			\bottomrule
		\end{tabular}
	}
\end{table}

\subsection{Baselines from Truth Discovery} 


The task in truth discovery is following: it assumes that sources vary in the trustworthiness, and one thus aims at learning the source reliability \citep[\eg,][]{Xiao.2015,Yao.2018}. As such, we can adapt it to our setting in that some entities (here: restaurants) vary in their level of trustworthiness, so that we learn an entity-individual quality score. Besides that similarity, there are also differences. Algorithms for truth discovery use only a univariate value as input, while multivariate input is absent. As such, other covariates such as, \eg, helpfulness or elite states are ignored.  

Truth discovery has some applications in a rating context. The typical application of it is to combine ratings from different platforms (Yelp, TripAdvisor, Google Maps, etc.). Yet this setting differs from rating aggregation: it is characterized by only \emph{few} sources $i\in\lbrace1,\ldots,N\rbrace$ but each with \emph{many} ratings $n_i$, \ie, $N \ll n_i$. In contrast, rating aggregation has \emph{many} sources (\ie, raters), yet each only \emph{few} scores, \ie, the per-source frequency of ratings is low. Due to this data scarcity, algorithms for inferences of rater-individual scores are mathematically not feasible in the setting of rating aggregation.  

(2) Truth discovery: truth discovery assumes that some entities (here: restaurants) vary in their level of “trustworthy” and, therefore, learns a source-individual quality measure (Xiao et al. 2015, Yao et al. 2018). 

Instead, they rely on rater-individual parameters which can not estimated for the same reason that the use of methods proposed for truth discovery \citep{Xiao.2015,Yao.2018} is precluded: they require a cross-sectional dataset, yet most raters provide only a single review, and, hence, an estimation of their reliability is prohibited.  This shortcoming is addressed by two hierarchical Bayesian models by pooling the rater-individual parameters \citep{Paun.2018} for which we also report the results. However, these models also fair poorly in comparison to the sample mean.


We experimented with different algorithms for truth discovery \citep{Xiao.2015,Yao.2018}, yet these were not applicable: truth discovery assumes that some raters are not trustworthy and, therefore, learns a source-individual reliability measure. Again, these parameters can not be learned based on only single observation from each rater.

\section{Source Code}
\label{appendix:source_code}

In the following, the model implementation in Stan \citep{Carpenter.2017} is provided. 

\SingleSpacedXI
\begin{lstlisting}[
  language=Stan,
%	frame=true,
  tabsize=2,
  showtabs=false,
  showspaces=false,	
  basicstyle=\scriptsize\ttfamily,
	keywordstyle=\bfseries\color{green!40!black},
%  commentstyle=\itshape\color{purple!40!black},
  commentstyle=\sffamily\itshape\color{white!40!black},
  identifierstyle=\color{blue},
  stringstyle=\color{orange},
	breaklines=true,
  numbers=left,
  numberstyle=\tiny\sffamily, 	
  firstnumber=1,
  stepnumber=5,
  numberfirstline=false,
	breaklines=true,
  breakatwhitespace=true,	
  prebreak=\mbox{\,$\color{darkgray}\mathbf{\hookleftarrow}$},	
]
data {
  int<lower=1> NUM_BUSINESS;//Number of businesses
  int<lower=1> TEST_NUM; //Number of ratings to be predicted
  int<lower=1> NUM_REVIEWS; // Number of observations/reviews
  int<lower=1> N[NUM_BUSINESS];//Number of ratings for each business
  int<lower=1,upper=5> Y[NUM_REVIEWS]; // Number of reviews
  int<lower=1> NUM_MEAN;
  real X[NUM_REVIEWS]; // Observation date (scaled)
  real X_test[NUM_BUSINESS*TEST_NUM]; //Out-of-sample timing
  matrix[NUM_REVIEWS,NUM_MEAN] Q_MEAN;
  matrix[NUM_MEAN,NUM_MEAN] R_MEAN;
  matrix[NUM_BUSINESS*TEST_NUM,NUM_MEAN] X_OOS_MEAN;
  real maxDiff[NUM_BUSINESS]; //Maximal difference between two ratings
  real minDiff[NUM_BUSINESS];//Minimal difference between two ratings
}
transformed data{
  vector[2] rho_prior[NUM_BUSINESS];
  real jitter = 1e-6;
  vector[5] counts=rep_vector(1,5);
  {
    for(j in 1:NUM_BUSINESS){  
      real mu = (minDiff[j] + maxDiff[j] + 1.0)/2.0;
      real sig = (maxDiff[j] - minDiff[j])/6.0;
      rho_prior[j,1] = 2.0 + mu^2/sig^2; 
      rho_prior[j,2] = (rho_prior[j,1] - 1) * mu;
    }
  }
}
parameters {
  vector<lower=0>[NUM_BUSINESS] rho; //Lengthscale paramter of exponential kernel
  vector<lower=0>[NUM_BUSINESS] sigma_GP;//Variance paramter of exponential kernel
  vector[NUM_REVIEWS] eta_raw;// Raw latent GP -- We use whitening, i.e. eta = L*eta_raw => eta ~ N(0,LL^T)
  vector[NUM_BUSINESS] R2;
  simplex[5] pi[NUM_BUSINESS];
  vector[NUM_MEAN] omega_tilde;
}
model {
  int idx = 1;
  vector[NUM_REVIEWS] MEAN = Q_MEAN * omega_tilde;
  eta_raw ~ std_normal();
  sigma_GP ~ std_normal();
  omega_tilde ~ std_normal();
  R2 ~ normal(0,5);
  for(j in 1:NUM_BUSINESS){
    matrix[N[j],N[j]] Cov;
    matrix[N[j],N[j]] L;
    vector[N[j]] eta;
    vector[4] cutpoints;
    real sq_sigma = square(sigma_GP[j]);
    pi[j] ~ dirichlet(counts);
    rho[j] ~  inv_gamma(rho_prior[j,1] , rho_prior[j,2]);
    cutpoints = inv_sqrt(1-inv_logit(R2[j]))*inv_Phi(cumulative_sum(pi[j,1:4]));
    for(i in 1:(N[j]-1)){
        Cov[i,i] = sq_sigma + jitter;
        for(k in (i+1):N[j]){
          real temp = sq_sigma*exp(-fabs(X[idx+k-1] - X[idx+i-1])/rho[j]);
          Cov[i,k] = temp;
          Cov[k,i] = temp;
        }
    }
    Cov[N[j],N[j]] = sq_sigma + jitter;
    L = cholesky_decompose(Cov);
    eta = L * segment(eta_raw,idx,N[j]) + segment(MEAN,idx,N[j]);
    segment(Y,idx,N[j]) ~ ordered_probit(eta,cutpoints);
    idx = 1 + sum(N[1:j]);
  }
}
generated quantities{
  vector[NUM_BUSINESS*TEST_NUM] predictions;
  vector[NUM_BUSINESS*TEST_NUM] rng_ratings;
  vector[NUM_REVIEWS] log_lik;
  vector[NUM_REVIEWS] fitted_values;
  vector[NUM_MEAN] omega =  R_MEAN\omega_tilde;
  {
    int idx = 1;
    vector[NUM_REVIEWS] MEAN = Q_MEAN * omega_tilde;
    vector[NUM_BUSINESS*TEST_NUM] OOS_MEAN = X_OOS_MEAN * omega;
    for(j in 1:NUM_BUSINESS){
      matrix[N[j],N[j]] Cov; 
      matrix[N[j],N[j]] L;
      vector[N[j]] eta;
      matrix[N[j],TEST_NUM] XX_star;
      matrix[TEST_NUM,N[j]] L_inv_k;
      vector[TEST_NUM] eta_pred;
      vector[4] cutpoints = inv_sqrt(1-inv_logit(R2[j]))*inv_Phi(cumulative_sum(pi[j,1:4]));
      real sq_sigma = square(sigma_GP[j]);
      for(i in 1:(N[j]-1)){
        Cov[i,i] = sq_sigma + jitter;
        for(k in (i+1):N[j]){
          real temp = sq_sigma*exp(-fabs(X[idx+k-1] - X[idx+i-1])/rho[j]);
          Cov[i,k] = temp;
          Cov[k,i] = temp;
        }
        for(l in 1:TEST_NUM){
          XX_star[i,l] = sq_sigma*exp(-fabs(X_test[(j-1)*TEST_NUM+l]-X[idx+i-1])/rho[j]);
        }
      }
      Cov[N[j],N[j]] = sq_sigma + jitter;
      XX_star[N[j]] = sq_sigma*exp(-fabs(to_row_vector(X_test[((j-1)*TEST_NUM+1):j*TEST_NUM])-X[idx + N[j]-1])/rho[j]);
      L = cholesky_decompose(Cov);
      L_inv_k = mdivide_left_tri_low(L,XX_star)';
      eta_pred = L_inv_k * segment(eta_raw,idx,N[j]) + segment(OOS_MEAN,(j-1)*TEST_NUM+1,TEST_NUM) ;
      eta  =  L * segment(eta_raw,idx,N[j]) + segment(MEAN,idx,N[j]);
      for(m in 1:TEST_NUM){
        vector[5] theta;
        for(r in 1:5){
          theta[r] = log(r) + ordered_probit_lpmf(r|eta_pred[m],cutpoints);
        }
        predictions[(j-1)*TEST_NUM+m] = exp(log_sum_exp(theta));
        rng_ratings[(j-1)*TEST_NUM+m] = ordered_probit_rng(eta_pred[m],cutpoints);
        }
      for(n in 1:N[j]){
        vector[5] theta;
        for(r in 1:5){
          theta[r] = log(r) + ordered_probit_lpmf(r|eta[n],cutpoints);
        }
        fitted_values[idx+n-1] = exp(log_sum_exp(theta));
        log_lik[idx+n-1] = ordered_probit_lpmf(Y[idx+n-1]|eta[n],cutpoints);
      }
      idx = 1 + sum(N[1:j]);
      }
  }
}
\end{lstlisting}

\end{APPENDICES}

\end{document}